\documentclass[]{bytedance_seed}
\usepackage{wrapfig}

\usepackage{minitoc}
\usepackage{url}
\usepackage{amssymb}
\usepackage[utf8]{inputenc}
\usepackage{microtype}
\usepackage{booktabs}
\usepackage{pifont} 
\usepackage{multirow}
\usepackage{makecell}
\usepackage{paralist}
\usepackage{xspace}
\usepackage{color}
\usepackage{xcolor}
\usepackage{colortbl}
\usepackage{adjustbox}
\usepackage{hyperref} 
\usepackage[edges]{forest}
\usepackage{tikz} 
\usepackage{caption}
\usepackage{amsfonts}
\usepackage[toc,page,header]{appendix}
\usepackage[utf8]{inputenc}
\usepackage{bbm}
\usepackage{bm}
\usepackage{enumitem}

\definecolor{seedc}{RGB}{7, 92, 173}

\newcommand{\name}[1]{}
\newcommand{\hardware}[1]{}

\renewcommand{\paragraph}[1]{\vspace{0.1em}\noindent\textbf{#1}}

\title{End-to-End Dexterous Arm-Hand VLA Policies via Shared Autonomy: VR Teleoperation Augmented\\by Autonomous Hand VLA Policy for\\Efficient Data Collection}

\author[]{ByteDance Seed}

\contribution[]{
Full Author List in Contributions}

\abstract{
Dexterous manipulation with multi-finger robotic hands remains a significant challenge for general-purpose robots. Recent Vision-Language-Action (VLA) models show promise in learning flexible skills from human-guided demonstrations, yet their scalability is limited by the scarcity of high-quality training data. Existing real-robot data collection faces inherent constraints: fully manual teleoperation imposes excessive cognitive load on human operators, limiting session duration, while automated planning often produces stiff, unnatural motions and yields suboptimal data distribution for learning skillful manipulation.

To address this, we propose a \textit{Shared Autonomy} framework that partitions control along the macro-micro motion domains. A human operator guides the robot end-effector via intuitive VR teleoperation, while an autonomous DexGrasp-VLA policy, using real-time tactile and local visual feedback, assists fine-grained and force-adaptive hand control as a \textit{Copilot}. This division of labor significantly reduces human cognitive load and enables efficient collection of high-quality data of coordinated arm-hand demonstrations with minimum mental fatigue.

Using these demonstration data, we train an end-to-end VLA policy enhanced with a novel \textit{Arm-Hand Feature Enhancement} module. This architecture explicitly captures both the \textit{distinct} latent features of macro (arm) and micro (hand) movements and their \textit{shared representations}, resulting in smoother and more robust arm-hand coordination. Furthermore, our \textit{Corrective Teleoperation} system supports continuous policy improvement through human-in-the-loop failure recovery and data augmentation. Experiments show that the framework produces high-quality data with very low operator effort, and the resulting fine-tuning can effectively train policies that achieve around 90\% success rate across more than 50 diverse objects, including unseen instances. The system's effectiveness is validated through comprehensive evaluations and ablation studies, highlighting its potential for advancing dexterous robotic manipulation.
}

\date{30 September, 2025}
\correspondence{\email{cuiyu.0627@bytedance.com}}

\checkdata[Project Page]{\url{https://dexvla-seed.github.io/dex-vla}}

\begin{document}
\maketitle


\section{Introduction}
\label{sec:intro}
The goal of general-purpose robots is to achieve physical intelligence comparable to humans for complex tasks through flexible and diverse manipulation. Humanoid robots with dexterous hands offer promising potential for this goal because of the form factor of anthropomorphic design, which naturally blends into human-centric environments. However, a critical bottleneck remains to unlock their full potential: advanced dexterous manipulation~\cite{rajeswaran2017learning, chen2023bi, ye2025dex1b}. This capability requires not only precise control of the robotic arm's spatial motion but also the execution of delicate hand actions and, crucially, seamless coordination between the two. Currently, learning such a level of arm-hand coordination remains an open question.

Recent advancements in end-to-end learning, particularly Vision-Language-Action (VLA) models~\cite{zitkovich2023rt, kim2024openvla, black2410pi0, intelligence2504pi0, bjorck2025gr00t}, have demonstrated promising performance in dexterous manipulation. However, VLA training, as a data-driven paradigm, relies on large-scale and high-quality demonstration datasets~\cite{o2024open, khazatsky2024droid, bu2025agibot}. Common data collection methods predominantly rely on fully manual teleoperation~\cite{zhao2023learning, ding2024bunny}, where an operator needs to simultaneously control all the Degrees-of-Freedom (DoFs) of both the arms and dexterous hands. Such a teleoperation scheme imposes an excessive cognitive load, resulting in low efficiency and scalability. From our practice, the reasonable average operational time is around 30 minutes without incurring extra mental fatigue. Most realistically, for untrained personnel, even 20 minutes of continuous full-DoF teleoperation is tiring.

Alternative approaches used reinforcement learning to learn manipulation skills by robots themselves~\cite{chen2023bi,lin2025rlsim2real,patel2025real,triantafyllidis2023hybrid}, or leverage motion planning~\cite{pan2025omnimanip,curtis2025trust, duan2024manipulate-any, huang2024rekep} to produce data. Although demonstrations and data are generated automatically, they often require extensive manual engineering beforehand, and typically it is not trivial to design good controllers for multi-finger dexterous tasks, or the automated policies are likely to produce unnatural manipulation behaviors, such as unwanted grasping poses. From our empirical practice, while automated planning and trajectory optimization methods can systematically generate data with a broad range of state-action pairs and high diversity through controlled, parameterized randomization, they suffer from two critical shortcomings for learning skillful manipulation. First, the generated motions often lack the sleekness of human movement, and the generated robot trajectories appear stiff and are inefficient speed-wise. Second, more fundamentally, as for imitation-based VLA training, the data distribution produced by automation is shaped by the planner's own solvers, constraints, and parameterized randomization. Although this distribution is often quantitatively diverse, it is qualitatively and unavoidably misaligned with the target distribution required for specific tasks. It fails to capture the nuanced task-relevant ``tricks'', compared to the priors of human experts learned in a lifelong fashion. Consequently, policies trained on such automated data may exhibit suboptimal performance, and fail to capture the nuanced task-relevant strategies exhibited inherently in the high-quality curated data from human demonstrations.

\begin{figure}[t]
    \centering
    \includegraphics[width=1.0\textwidth]{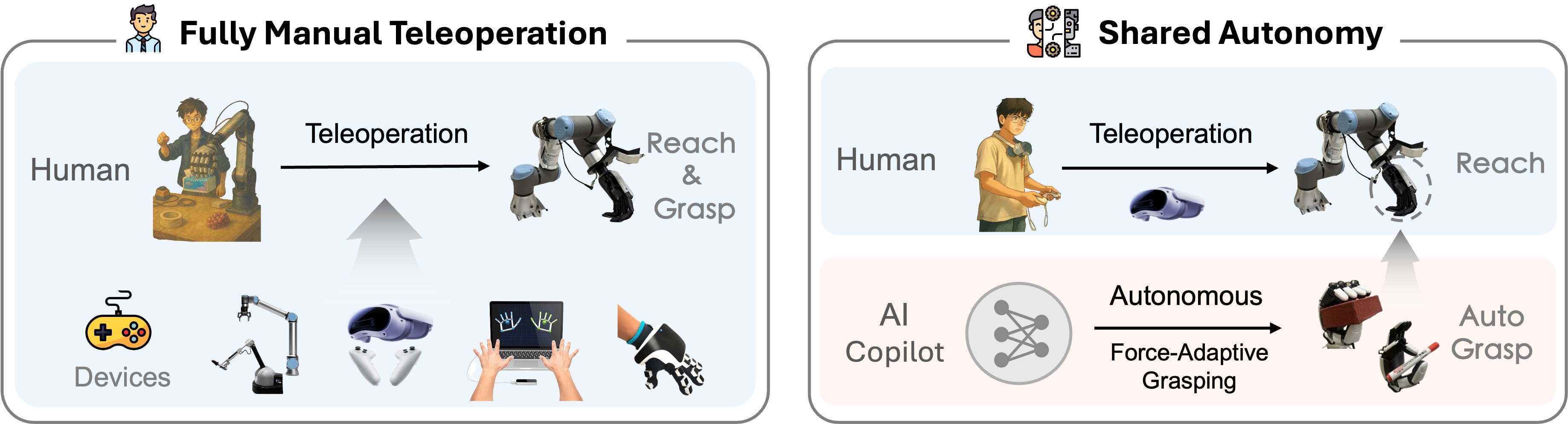}
    \caption{\normalsize Comparison of fully manual teleoperation and the proposed \textit{Shared Autonomy} -- semi-autonomous teleoperation, where the human operator focuses on high-level control of the arm motion, while the hand is controlled by the perceptual AI-Copilot for autonomous grasping. }
    \label{fig:copilot_paradiam}
    \vspace{-5mm}
\end{figure}

To address these limitations, we propose a \textit{Shared Autonomy} framework (Fig.~\ref{fig:copilot_paradiam}) that partially offloads the high-level arm teleoperation from fine-grained low-level autonomous control of multiple fingers of the dexterous hand, enabling efficient collection of high-quality and coordinated demonstrations with low mental load. The core innovation lies in a functional division of shared control that leverages the complementary strengths of humans and AI (the autonomous VLA policy for the hand): the human focuses on high-level scene understanding, semantics, spatial reasoning, and affordance to teleoperate the robot end-effector easily via the Virtual Reality (VR) interface. Meanwhile, a learned DexGrasp-VLA controller acts as a VLA Copilot that autonomously executes fine-grained and compliant grasping motions with the dexterous hand. The DexGrasp-VLA copilot for the hand is a tactile-based VLA policy for the dexterous hand. It integrates language commands, in-hand camera images, proprioceptive states, and two complementary tactile features: resultant force vectors (magnitude and direction) and spatial tactile embeddings (contact distribution patterns). This multi-modal perception of the DexGrasp-VLA provides sufficient feedback for robust and adaptive grasping performance across diverse conditions. 

This shared-autonomy paradigm eliminates the need for the operator to simultaneously control all degrees of freedom of a complex arm-hand system. By delegating high-DoF hand control to the VLA Copilot, we significantly reduce the cognitive load from the human operator, produce naturally coordinated arm-hand trajectories with stable grasps, and enable faster data collection with a high success rate. It shall be noted that this work uses VLA policies in two different ways: one is the hand-only tactile-aware VLA, and the other is for the holistic arm-hand control -- the DexGrasp-VLA is a sub-policy for controlling the dexterous hand within the full VLA framework for data collection and model training. 

Using these collected trajectories, we perform Supervised Fine-Tuning (SFT) of a pre-trained VLA model to learn an end-to-end policy for holistic arm-hand control. A key architectural innovation is our \textit{Arm-Hand Feature Enhancement} module, which explicitly models the complementary roles of the arm and hand. The module is comprised of two complementary features: (1) a shared task representation which is encoded originally by the base VLA model from visual, language, and state inputs; and (2) two additional separate encoders for the arm and the hand, respectively. Each branch of features is optimized under auxiliary losses that encourage specialization toward macro-movement (reaching) and micro-manipulation (grasping) capabilities. The resulting disentangled features are then combined with the original shared representation, allowing the model to retain the global task context while incorporating the limb-specific dynamics. 

This design is motivated by the fundamental kinematic and operational differences between arm and hand motions: the arm requires smooth, long-horizon control of the end-effector poses, while the hand requires compliant and contact-rich hand-object interactions. Experiments confirm that this structure leads to more efficient learning of coordination patterns, much more robust performance under visual disturbances (camera occlusions), and more natural emergent behaviors compared to a monolithic architecture that predicts all joint commands from a single undifferentiated representation.

Finally, we implement a \textit{Corrective Human-in-the-loop Teleoperation} to enable continuous self-improvement of the deployed policy. During execution of the model inference, successful trajectories are automatically recorded to form the positive demonstration datasets. More importantly, when the policy fails due to unseen object shapes, adversarial configurations, or environmental disturbances, the system allows a human operator to actively intervene in real time through the same shared autonomy interface. The operator takes over the control, corrects the failure, and completes the task, thereby producing new data of recovery trajectories. Both the successful and recovery data are aggregated into training datasets for iterative model refinement, enabling incremental improvements on the policy and thus gradually enhancing robustness, as well as expanding its capability to handle the real-world variability and corner cases.

The core technical contributions are as follows: 
\begin{itemize}
[leftmargin=*]
    \setlength{\itemsep}{1pt}
    \setlength{\parskip}{0pt}
    \setlength{\parsep}{0pt}
\item \textbf{Novel Multimodal VLA Copilot for Dexterous Grasping:} We present DexGrasp-VLA, the first Vision-Language-Action (VLA) Copilot that robustly fuses visual, tactile, language, and proprioceptive feedback for autonomous, force-adaptive grasping with a five-finger hand. Its reactive and compliant fine-grained grasping capability represents a major advance over prior data collection approaches which lacked consistent performance in contact handling and force adaptability.

\item \textbf{Shared Autonomy for Efficient Data Collection:} We propose a paradigm-shifting framework that strategically divides control between the human operator and the VLA Copilot to overcome the data bottleneck in dexterous manipulation. By combining intuitive human VR teleoperation for high-level arm guidance assisted with the tactile-enhanced DexGrasp-VLA Copilot for autonomous fine-grained grasping, we substantially reduce operator cognitive load and enable efficient collection of high-quality demonstration data.

\item \textbf{End-to-End VLA with Arm-Hand Feature Enhancement:} We design a novel VLA architecture that explicitly disentangles the control of macro (arm) and micro (hand) movements through dedicated feature pathways, while preserving a shared global task representation. This architectural innovation directly solves the problem of monolithic controllers that fail to capture the distinct kinematics and dynamics of the arm and hand respectively, thus leading to significantly more robust and generalizable arm-hand coordination, with notably improved resilience to visual occlusions.

\item \textbf{Corrective Human-in-the-Loop Teleoperation for Continuous Policy Improvement:} We introduce a corrective human-in-the-loop intervention that operates without interrupting deployed policy execution, enabling purposeful data collection from long-tail failure cases during real-world deployment. Such teleoperation allows humans to assist imperfect VLA executions, producing complementary datasets for scenarios that would otherwise fail. Integrating these curated recovery data with successful trajectories, we establish a self-improving data flywheel: retraining the base VLA model on the iteratively augmented dataset results in an updated policy with enhanced robustness in real-world applications. This approach extends the training by unifying both pre-scripted demonstrations with in-situ deployment data, thereby enabling targeted coverage of corner cases that are otherwise impractical to anticipate beforehand.
\end{itemize}

Our experimental results demonstrate the effectiveness of the proposed framework. It enables the efficient data collection of high-quality and coordinated arm-hand demonstrations, which in turn underpins the learning of an end-to-end VLA policy that achieves a 90\% success rate across a diverse set of over 50 objects, including unseen objects and settings. The necessity of each core component was systematically validated through comprehensive ablation studies, which confirmed that the DexGrasp-VLA model, the Arm-Hand Feature Enhancement module, and the Corrective Teleoperation system are all essential and significantly improve the policy's performance, success rate, and robustness. These strong empirical results provide an essential testbed for exploring broader algorithmic strategies before scaling up to larger models.

This work is strategically positioned within the broader challenge of developing robust foundational models for embodied intelligence with a focus on dexterous manipulation. While scaling laws have achieved remarkable progress in large vision and language models, the direct application to robotics still faces significant data quality and architectural challenges, due to the fact that motion trajectories, visual streams, and high-frequency tactile-force signals are fundamentally heterogeneous in nature. Simply brute-force directing heterogeneous data into a monolithic architecture is unlikely to yield the desired emergent dexterity, or effectively leverage scaling laws in a scientific manner. 

Our work argues that a necessary premise to effective large-scale pre-training is the foundational understanding of how to algorithmically integrate these distinct sensorimotor pathways. By leveraging SFT of pre-trained VLA models, we agilely iterate research investigation: we can rapidly isolate and validate core components, such as our \textit{Arm-Hand Feature Enhancement} module and effective tactile integration strategies, within a controlled setting. This approach efficiently exploits the pre-trained models' semantic and generalization capabilities, and meanwhile allowing us to focus on the fundamental physical principles of useful multimodal fusion. Thus, the effective algorithms developed and validated on a smaller scale provide the essential blueprint for the future collection of massive, high-quality datasets (e.g., via our Shared Autonomy framework), and strengthen higher likelihood of success in the subsequent pre-training of more sophisticated large-scale robot foundational models with true human-like physical intelligence.

\section{Related Work}
\label{sec:related_work}
\subsection{Vision-Language-Action Models for Robot Control}

Recent advances in Vision-Language-Action (VLA) models represent a significant step change toward general-purpose robots that interpret natural language instructions and visual observations for downstream control tasks~\cite{zitkovich2023rt, kim2024openvla, black2410pi0, intelligence2504pi0, bjorck2025gr00t}. These models typically leverage large-scale pre-trained vision-language models (VLMs) as semantic backbones and fine-tune lightweight action heads to map multi-modal inputs into robot control commands. Most existing works, however, focus on low-degree-of-freedom (DoF) end-effectors, such as two-finger grippers or parallel jaw grippers~\cite{liu2025hybridvla, deng2025graspvla, wen2025dexvla}, and are often evaluated on short-horizon tasks like pick-and-place. This limits their applicability to real-world manipulation scenarios which demand rich contacts and high dexterity.

An increasing amount of research work has been investigating how to extend the VLA frameworks to dexterous hand manipulation. The work in~\cite{zhong2025dexgraspvla} proposes a DexGraspVLA model as a hierarchical architecture that combines a pre-trained vision-language planner with a diffusion-based low-level action controller. It transforms diverse language and visual inputs into domain-invariant representations, enabling robust imitation learning. Being-H0~\cite{luo2025being} takes a data-centric approach, pretraining VLA models on large-scale human hand videos with a novel physical instruction tuning paradigm. The model utilizes part-level motion tokenization for millimeter-accurate trajectory learning and 3D physical space alignment, thereby enhancing generalization. In parallel, other research has been conducted to incorporate tactile sensing into VLA models. For example, OmniVTLA~\cite{cheng2025omnivtla} introduces a Vision-Tactile-Language-Action architecture with a dual-path tactile encoder, consisting of a vision transformer and a semantically-aligned tactile ViT. 

Although these methods demonstrate strong performance, they generally treat the arm–hand system as a single monolithic controller, overlooking the fundamental kinematic and dynamic differences between the arm and the hand: the arm serves as a floating base for long-range positioning and pre-grasp interactions, while the hand performs fine-grained and contact-rich multi-finger interactions to acquire various objects. 

Based on the first principle of the different characteristics in motion and physical interactions between the hand and arm, a monolithic design in VLA leads to two key limitations. First, training the arm-hand policies in a monolithic manner fails to capture the distinct kinematic and dynamic properties of macro-level arm motion versus micro-level hand manipulation, potentially limiting coordination and robustness. Second, these systems rely heavily on large-scale, offline-collected demonstrations, either through scripted control policies or direct full-DoF teleoperation \cite{mao2024dexskills} with exoskeletons, which are costly and difficult to scale, creating a data bottleneck especially for dexterous hands. Moreover, these systems often employ an undifferentiated fusion of tactile features, lacking explicit representation of contact information for more precise action generation of the VLAs to fine control hand motions.

In terms of the technical differentiation that overcomes these limitations, our work proposed a novel \textit{Shared Autonomy} framework that decouples the control problem to enable tractable data collection. The human only needs to teleoperate the arm via a VR interface, while the VLA Copilot autonomously controls the multi-finger hand using a tactile-enhanced VLA policy. Further, to complement this, we developed an \textit{Arm–Hand Feature Enhancement} architecture that explicitly disentangles shared semantic representations into limb-specific embeddings for the arm and hand respectively. This structured representation is then augmented by the original shared arm-hand features, generating more granular control signals for natural, fine-grained, and robust arm-hand coordination. This co-design of a scalable data collection and a structured policy architecture improves interpretability, modularity, and generalization in dexterous manipulation tasks.

\subsection{Existing Data Collection Approaches}

High-quality demonstrations are critical for imitation learning, especially in high-DoF, contact-rich dexterous manipulation. Teleoperation-based methods, such as leader-follower setups~\cite{zhao2023learning}, vision-based retargeting~\cite{qin2023anyteleop}, and VR interfaces~\cite{cheng2024open, zhao2025xrobotoolkit}, can produce precise, human-like trajectories but are difficult to scale due to the excessive reliance on \textit{skilled human operators}. On the other hand, automated pipelines that use motion planners~\cite{chen2023bi, lin2025rlsim2real, patel2025real, thomason2024motions, kumar2024open}, including CuRobo~\cite{sundaralingam2023curobo}, enable efficient large-scale data generation but often lack contact sensitivity, self-correction ability, or contextual adaptability that are inherent in human demonstrations.

Recent efforts have explored simulation-based data generation frameworks~\cite{mu2025robotwin, chen2025robotwin2, geng2025roboverse, nasiriany2024robocasa}, which can produce large-scale synthetic demonstrations. While simulation provides a scalable alternative for pre-training base models, real-world data remains essential for sim-to-real transfer, fine-tuning, and co-training. As emphasized in prior work~\cite{maddukuri2025simreal}, even with sophisticated sim-to-real techniques, co-training with real robot data is necessary to bridge the reality gap. Importantly, our preliminary findings show that the amount of real-world data required for effective sim-to-real co-training is comparable to that needed for direct real-world fine-tuning, which is typically on the order of hundreds of demonstrations for dexterous manipulation tasks (at least at the scale of several dozen demonstrations for training task-specific narrow skills).

Motivated to address these limitations, we developed Shared Autonomy with VLA-Copilot that combines human high-level arm guidance with autonomous hand-level control using tactile feedback. This design reduces cognitive load, captures both human intent and reactive adaptation, and enables efficient collection of high-quality real-world demonstrations, which are crucial for training generalizable VLA policies.

\subsection{Tactile Sensing for Robot Manipulation}

Tactile sensing provides essential feedback for capturing contact dynamics during physical interactions, especially in tasks involving compliant interaction, precision, or uncertainty. Prior works have explored vision-tactile fusion in reinforcement learning, demonstrating effectiveness in assembly~\cite{hansen2022visuotactile, lee2020making} and dexterous in-hand manipulation~\cite{liuvtdexmanip, hu2025vt-dexterous}. These methods typically use tactile signals as feedback for closed-loop policies in contact-rich scenarios.
More recently, studies applied imitation learning to learn combined vision-tactile representations for fine-grained control~\cite{xue2025reactive-vtac, liu2025vitamin, huang20243dvitac, lin2024visuotactile}, enabling reactive manipulation with end-to-end learning from multimodal demonstrations.

Building on these, recent methods begin to incorporate tactile perception into Vision-Language-Action (VLA) frameworks~\cite{bi2025vla-touch, huang2025tactile-vla, cheng2025omnivtla, zhang2025vtla}, improving generalization and robustness across object categories and tasks.
For example, VLA-Touch~\cite{bi2025vla-touch} enhances a frozen VLA backbone with dual-level tactile feedback: a pretrained tactile-language model assists in high-level semantic planning, and a diffusion controller refines low-level execution.
Tactile-VLA~\cite{huang2025tactile-vla} further integrates tactile signals into planning and control, combining hybrid position–force control with commonsense reasoning based on vision-language priors. This design enables zero-shot generalization using minimal tactile supervision.
VTLA~\cite{zhang2025vtla} focuses on fine contact tasks such as insertion, proposing a Vision-Tactile-Language model trained via Direct Preference Optimization, achieving high success in both simulation and real-world peg-in-hole tasks.
OmniVTLA~\cite{cheng2025omnivtla} adopts a data-centric approach, constructing a large tri-modal dataset and learning semantically aligned tactile embeddings across diverse sensors, improving performance for both grippers and dexterous hands. 

Despite nuanced differences in technical details, these methods all attempt to fuse tactile inputs into the VLA backbone, through full end-to-end retraining, multimodal pretraining, or task-specific architectural improvements. Such strategies often rely on large-scale tactile datasets, or require architectural changes that add significant complexity to the model. In contrast, our method uses a modular design approach: we inject tactile feedback locally into the action expert to enrich the contact-relevant information for improving contact-rich control, while keeping the vision-language backbone totally intact, adding no extra complexity. This allows for efficient learning and deployment without full retraining while preserving useful tactile grounding during execution.

\section{Methods and Training Recipe}
\label{sec:data_training}

\begin{figure*}[t]
     \centering
     \includegraphics[width=\textwidth]{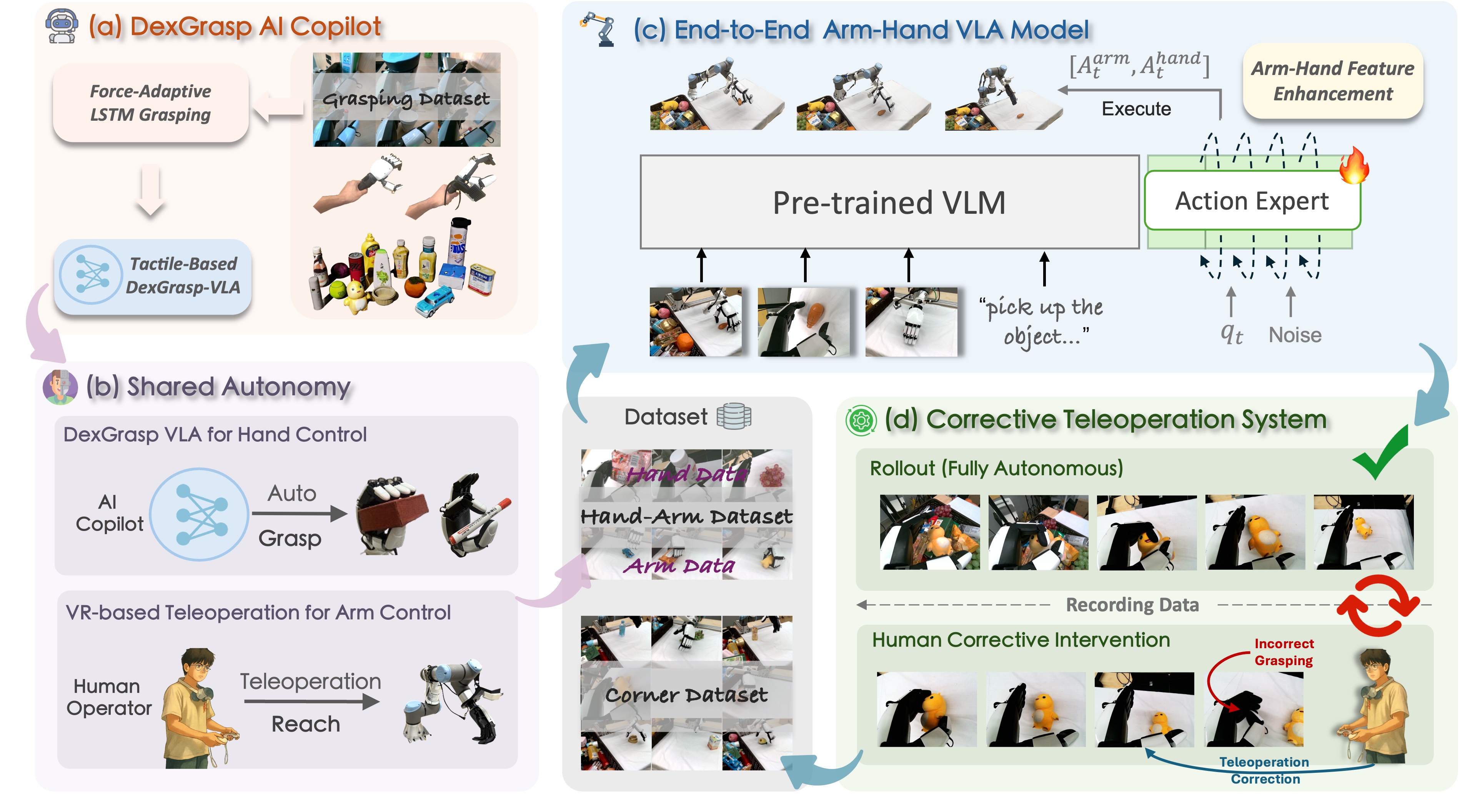}
     \vspace{-5mm}
     \caption{Data collection and training pipeline for DexGrasp-VLA policy and arm-hand VLA policies: (a) Tactile-based DexGrasp-VLA policy for a five-finger dexterous hand; (b) Shared autonomy data collection; (c) End-to-end arm-hand policy learning with Arm-Hand Feature Enhancement; (d) Corrective human-in-the-loop teleoperation.}
     \label{fig:title-teaser}
     \vspace{-3mm}
 \end{figure*}

We present an integrated pipeline for learning dexterous arm-hand policies, which trains the control of the dexterous hand and arm coherently, as illustrated in Fig.~\ref{fig:title-teaser}. The pipeline consists of four key stages:
\begin{itemize}
[leftmargin=*]
    \setlength{\itemsep}{1pt}
    \setlength{\parskip}{0pt}
    \setlength{\parsep}{0pt}
\item First, we train the DexGrasp-VLA controller to perform autonomous grasping using multimodal feedback for the dexterous hand (visual and tactile sensing). This controller serves as the core VLA Copilot in our shared autonomy framework, providing robust force-adaptive grasping capabilities.

\item Second, within the shared autonomy framework, a human operator teleoperates the arm via a VR interface, while the DexGrasp-VLA acts as a VLA Copilot for controlling the hand. This division of labor enables efficient collection of synchronized arm-hand data with significantly reduced cognitive load for the operator.

\item Third, using such synchronized data, we train a holistic end-to-end policy augmented with our novel arm-hand feature enhancement module. This architecture extracts and fuses dedicated features specifically for the arm and hand, which significantly improves the coordination between macro- and micro- movements.

\item Finally, we implement a corrective teleoperation system that continuously improves the policy by incorporating both successful trajectory data and human-guided recovery demonstrations. This iterative refinement process enhances the robustness of the policy against edge cases and distribution shifts.
\end{itemize}

Together, these components form a complete system for acquiring and continuously refining dexterous manipulation skills through efficient data collection and targeted policy improvement.

\subsection{Design Philosophy and Problem Formulation}

The design philosophy of this work stems from the principle of a collaborative human-AI framework. Similar to the work in \cite{aisearch2025} for bronchoscopic foreign body removal, where an AI-Copilot robot autonomously searches for foreign bodies, and a physician remotely performs the extraction via teleoperation -- such synergy leverages AI for simpler search tasks for more efficient explorations, and human expertise for more difficult and delicate removal, enhancing precision and safety. Likewise, for grasping tasks involving pinch-grasp, power-grasp, and their derived grasping synergies -- as those studied in this paper which are suitable for logistic applications -- the visual-tactile based grasping can be autonomously executed by VLA-based hand control, while humans can perform guidance of a diversity of pre-grasp interactions and positioning, which are much more efficient during the data collection stage. 

Following the coherent collection of synchronized data, our ultimate goal is to learn an end-to-end VLA model capable of controlling all the joints coherently for coordinated arm-hand grasping. Both DexGrasp-VLA policy for the hand, and the end-to-end arm-hand VLA policy for the hand and arm, are fine-tuned from $\pi_0$~\cite{black2410pi0} using the open-source framework LeRobot~\cite{cadene2024lerobot}. 

Specifically, the model predicts a sequence of future actions conditioned on the current observation. Let $\mathcal{A}_t = [a_t, a_{t+1}, \dots, a_{t+H-1}]$ denote a horizon-$H$ action sequence. The model aims to approximate the conditional distribution $\pi(\mathcal{A}_t \mid o_t)$, where $o_t$ is the observation at time $t$. Each observation $o_t$ includes multiple RGB views, a language command, and proprioceptive states: $o_t = [I^1_t, \dots, I^m_t, l, q_t]$. Here, $I^i_t$ represents the $i$-th camera image, $l$ is the tokenized instruction, and $q_t$ is the robot’s joint state vector. All modalities are encoded through modality-specific encoders and projected into a shared embedding space for cross-modal reasoning and control.


\subsection{DexGrasp-VLA: Autonomous Dexterous Grasping Policy}

Our shared autonomy framework builds on DexGrasp-VLA, a high-performance controller that acts as a VLA copilot for dexterous hand grasping: it autonomously executes fine-grained manipulation through multimodal sensory feedback, relieving the human operator from control of the hand's \textit{12 DoFs} and allowing them to focus exclusively on arm motion (for 21-DoF hand with higher dexterity, direct teleoperation would be even harder). 
To ensure robustness and generalization, DexGrasp-VLA is developed through a two-stage training pipeline. It begins by learning a hand-only LSTM policy from a hybrid dataset consisting of parameterized force-control and teleoperated demonstrations. This compact, blind policy captures rich contact behaviors using tactile sensing and enables force-adaptive grasping, allowing the dexterous hand to close quickly and then tighten gradually for a stable grasp. The DexGrasp-VLA itself acts autonomously and is used to collect various grasping data. Building upon this, we train a hand-centric VLA policy that further integrates visual and tactile sensing, enabling perceptual grasping that is context-aware and reactive.

\subsubsection{\textbf{Force-Adaptive Grasping Policy Learned by LSTM}} 
\label{ssec:lstm-policy}

\begin{figure}[!h]
    \centering
    \vspace{-3mm}
    \includegraphics[width=1.0\textwidth]{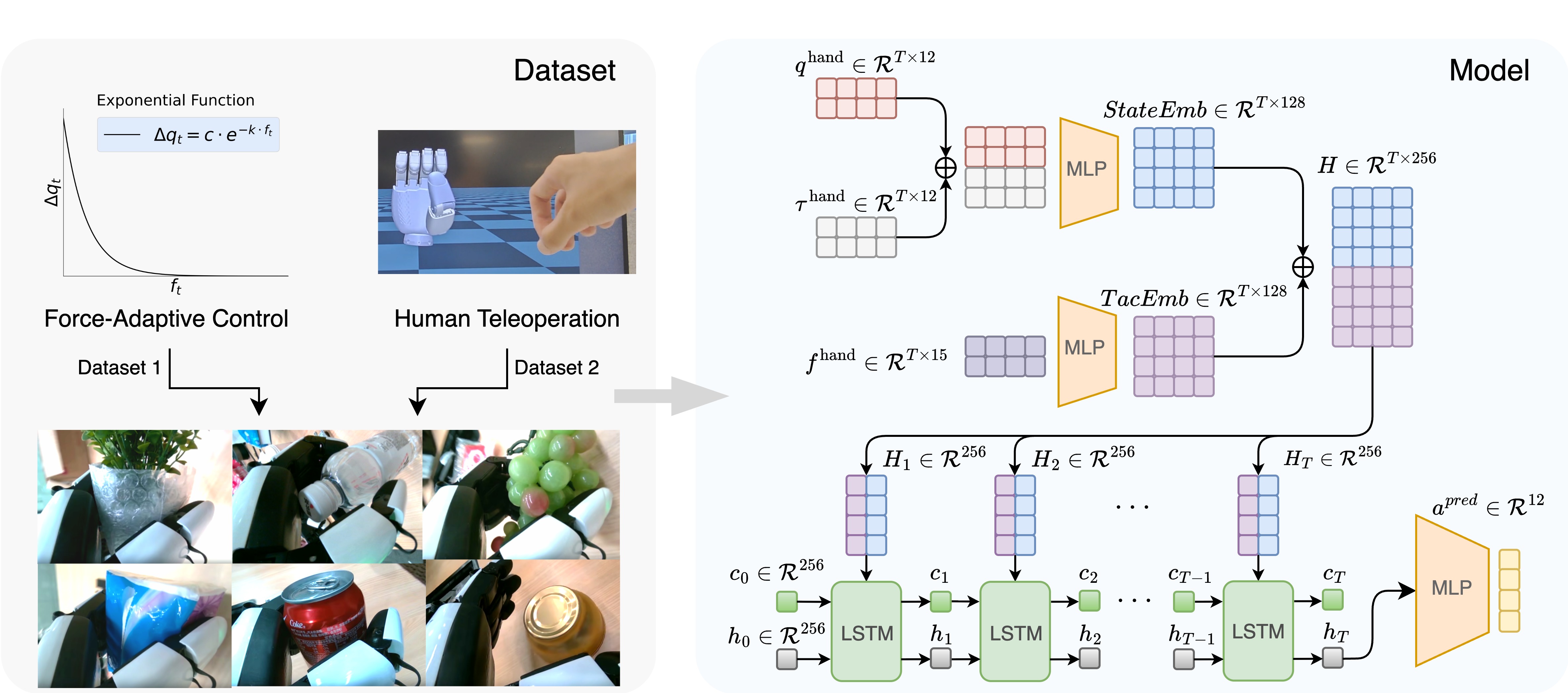}
    \caption{\normalsize Data collection and training framework for the force-adaptive grasping policy learned via LSTM. The datasets are collected by parameterized control (68 demos) and teleoperated human strategies (150 demos), respectively: the former captures the implicit tactile/force-sensitive grasping, and the latter incorporates more flexible multi-finger form closure from human teleoperation, ensuring data diversity. }
    \label{fig:lstm}
\end{figure}

To bootstrap reactive grasping under rich contact dynamics without complex visual perception, we first train a ``blind'' policy \cite{wang2022learning} using solely the proprioceptive and \textit{tactile feedback}. The policy uses a hybrid dataset from a parametric force-based controller (68 demos) and human teleoperation demonstrations (150 demos), ensuring force-sensitive compliance control and human-like grasping strategies. We use a diverse set of over 20 objects with varying shapes, sizes, weights, and materials to enhance the \textit{implicit} behavior of variable adaptive grip forces, given the proprioceptive–tactile history encapsulated by the LSTM. 

\paragraph{\textbf{Data Collection for LSTM}}

Data are collected through two complementary methods:
1) \textit{Force-Adaptive Position Control:}
A rule-based controller generates compliant grasping at 50 Hz. At each step \(i\), the command \(q_c(i)\) is computed as:
\begin{equation} \label{eq:force-control}
q_c(i) = q_m(i) + q(0) \cdot e^{-k \cdot f_{z}(i)},
\end{equation}
where \(q_m(i)\) is the measured joint position, \(q(0)\) is the initial position, \(k\) is the gain to adjust the grasping force , and \(f_z(i)\) is the fingertip resultant normal force. This control law enables rapid hand closure when no contact is detected, and gradually increases the grip force upon contact. 

2) \textit{Human Teleoperation via Retargeting:}
A \textit{Leap Motion} sensor is used to capture human hand motions, retargeted to the 12-DoF \texttt{Xhand}. Each demonstration covers the full grasping sequence from hand opening to stable grasp, sampled at 50 Hz.

\paragraph{\textbf{Training LSTM-based Policy via Behavior Cloning}}

As shown in Fig.~\ref{fig:lstm}, unifying the diverse behaviors from automated and teleoperation-based data, we train a lightweight LSTM-based policy that fuses these grasping strategies into a compact state-based controller. This design achieves three objectives: (i) computational efficiency for real-time onboard deployment, (ii) generalization across diverse objects and contact conditions by capturing temporal dependencies in tactile-proprioceptive history, and (iii) scalable autonomous data generation without continuous human supervision. Once trained, this policy serves as a unified controller bridging heterogeneous data sources and a practical control expert to bootstrap training of vision-based VLA policies.

We implement through behavior cloning, where the input at each time step $t$ is represented as $x_t = [s_t^{\text{hand}}, f_t^{\text{hand}}] \in \mathbb{R}^{39}$, with $s_t = [q_t^{\text{hand}}, \tau_t^{\text{hand}}] \in \mathbb{R}^{24}$ denoting the proprioceptive state, and $f_t^{\text{hand}} \in \mathbb{R}^{15}$ representing signals measured from fingertip tactile arrays. A sliding window of length $T$ forms the sequence input $X = [x_{t-T+1}, \dots, x_t] \in \mathbb{R}^{T \times 39}$, ensuring temporal continuity.  

The policy network uses two parallel three-layer MLPs with ReLU and batch normalization to encode proprioceptive and tactile inputs independently. Outputs are concatenated into a fused feature representation $H \in \mathbb{R}^{T \times 256}$, passed into a single-layer LSTM (hidden layer size of 256) to capture essential temporal dependencies for contact-rich interactions. The output is mapped through a fully connected layer to predict the hand action $a_t \in \mathbb{R}^{12}$.  

The model is optimized with a mean squared error (MSE) loss between predicted and demonstrated actions, plus a $\mathcal{L}_{2}$ regularization:
\begin{equation} \label{eq:training-loss}
 \mathcal{L} = \frac{1}{N} \sum_{i=1}^N \left\| a_t^{(i)} - \hat{a}_t^{(i)} \right\|_2^2 + \lambda \|\Theta\|_2^2,
\end{equation}  
where N denotes the number of samples, and the second term represents the regularization penalty.

The trained LSTM policy functions as a vision-free reactive expert, producing robust grasping trajectories from tactile feedback. Its main role is to make large-scale autonomous data collection easier, thereby enriching data diversity and improving VLA policy generalization. It should be noted that different object properties such as stiffness are indirectly captured by proprioceptive–tactile history; diverse multi-finger closures stem from human demos, while stable grip force and compliance control (without damaging or crushing objects) are provided by the force-control method, where the settle-down force is implicitly learned by behavior cloning in an admittance manner (Equation \ref{eq:force-control}). Future dexterous hands with advanced force control could achieve this behavior directly by controlling reference contact forces for different objects, offering higher performance.

\subsubsection{\textbf{Tactile-based DexGrasp-VLA $\pi_{\text{hand}}$}}
\begin{figure}[!h]
    \centering
    \includegraphics[width=1.0\textwidth]{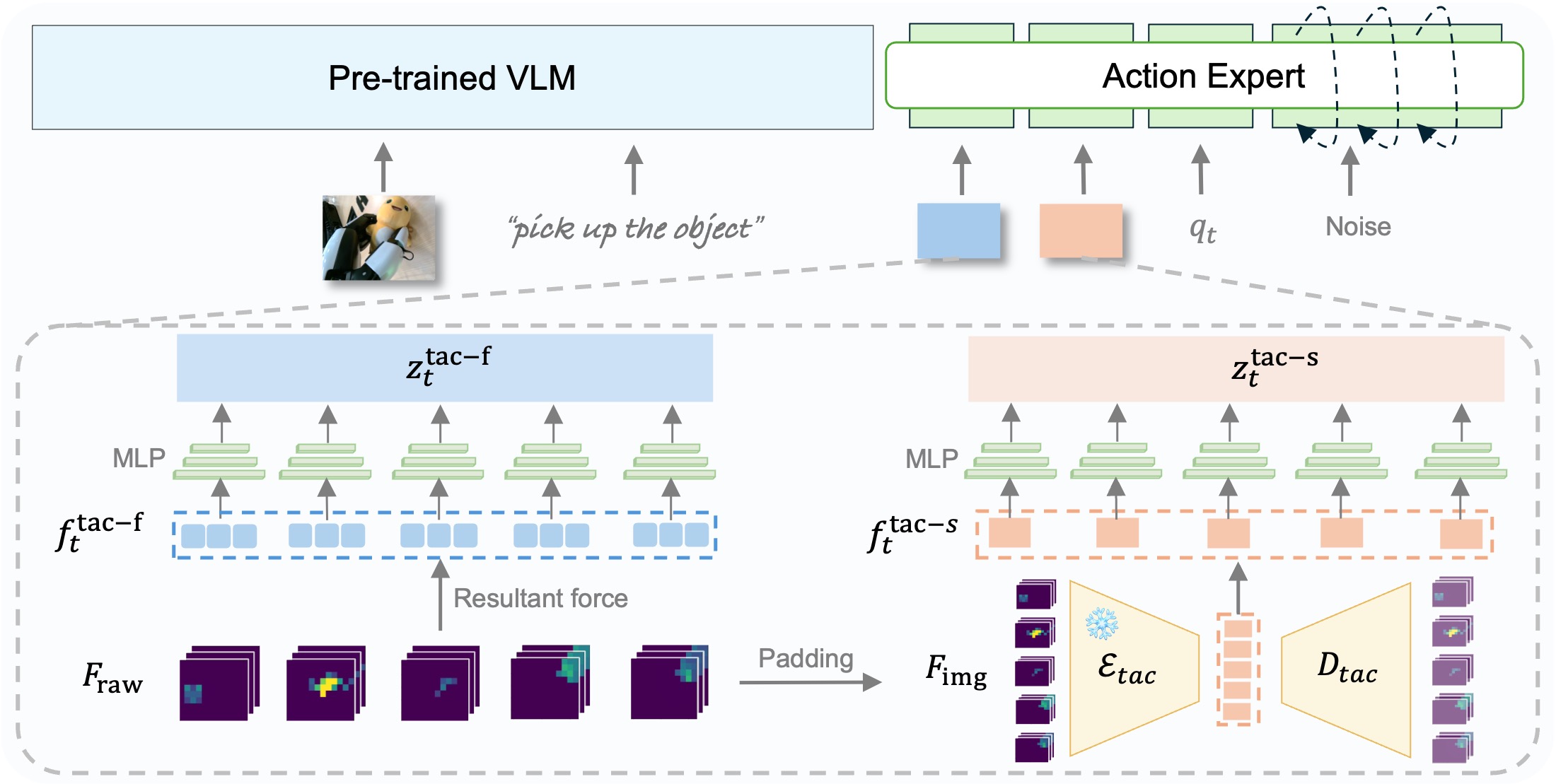}
    \caption{\normalsize Tactile-based DexGrasp-VLA for hand control. Two complementary tactile features are extracted: (i) resultant forces, representing the net contact force vector at each fingertip $z_t^{\text{tac-f}}$, and (ii) tactile image latents $z_t^{\text{tac-s}}$, capturing spatial contact patterns through a convolutional autoencoder.}
    \label{fig:tac_vla}
\end{figure}

Although the LSTM policy provides robust low-level force adaptation, it lacks visual perception and scene understanding. To incorporate visual context and enable task-aware grasping (e.g., timing to grasp the target object in clutter), we use the data autonomously collected by the LSTM policy to train a multimodal VLA policy that integrates tactile sensing for robust dexterous grasping, as detailed in Fig.~\ref{fig:tac_vla}.

\paragraph{\textbf{Tactile Feature Extraction}}

The raw tactile readings from the dexterous hand sensors are captured as a high-dimensional tensor \( F_{\text{raw}} \in \mathbb{R}^{10 \times 12 \times 3} \) per fingertip, representing force values along the \(x\), \(y\), and \(z\) axes of a spatial grid over the surface of the fingertip. Directly feeding such high-dimensional data into a VLA policy is computationally inefficient. To overcome this, we derive \textit{two complementary tactile features} that provide a compact yet rich representation of contact conditions, as shown in Fig.~\ref{fig:tac_vla}.

The first feature, the resultant force vector \( f_t^{\text{tac-f}} \in \mathbb{R}^{5 \times 3} \), is obtained by summing force components at each fingertip’s sensor array. This offers a physically interpretable and explicit measurement of the magnitude and direction of the net contact force per fingertip. However, this representation lacks spatial detail about distribution of contact forces.

To capture spatially distributed contact forces, contact patterns and shear forces, we introduce a complementary tactile feature $f_t^{\text{tac-s}} \in \mathbb{R}^{5 \times 128}$ using a convolutional autoencoder (CAE). This approach encodes the spatial-force relationships across each fingertip into a compact latent representation. Each raw tactile sensor array $F_{\text{raw}}$ is normalized and zero-padded to form a $16 \times 16 \times 3$ tactile image $F_{\text{img}}$, where 3 channels corresponding to the $x$, $y$, and $z$ force components.

The encoder comprises three convolutional layers: $3 \times 3$ kernels and a stride of 2, using 32, 64, and 128 filters, respectively. 
Each layer is followed by batch normalization and a ReLU activation, progressively reducing the spatial resolution while increasing feature depth. This hierarchical encoding results in a \textit{$2 \times 2 \times 128$ feature map}, which is flattened and projected to a \textit{128-dimensional latent vector} per fingertip. The decoder mirrors this architecture using transposed convolutions to reconstruct the original tactile image from the latent code.

The CAE is trained to minimize the reconstruction loss:
\begin{equation} \label{eq:recon-loss}
\mathcal{L}_{\text{recon}} = \frac{1}{3HW} \sum_{c \in \{x,y,z\}} \sum_{i=1}^{H} \sum_{j=1}^{W} \left( F_{c,ij} - \hat{F}_{c,ij} \right)^2,
\end{equation}
where $H = W = 16$. This reconstruction objective ensures that the latent representation preserves the essential spatial and directional characteristics of the contact force distribution.
When combined with the resultant force feature $f_t^{\text{tac-f}}$, this spatial tactile embedding provides a more explicit and detailed representation of the contact conditions, enabling more robust and adaptive grasping policies.

\paragraph{\textbf{Grasping VLA Policy Learning}}

To integrate the extracted tactile features into the VLA framework, we transform the per-fingertip force representations -- the resultant force vector \( f_t^{\text{tac-f}} \) and the spatial tactile embedding \( f_t^{\text{tac-s}} \) -- into embedding vectors \( z_t^{\text{tac-f}} \) and \( z_t^{\text{tac-s}} \) via separate multi-layer perceptrons (MLPs), as shown in Fig.~\ref{fig:tac_vla}. These tactile embeddings are then fused with pre-processed embeddings from other sensory modalities, including the visual feed from the eye-in-hand camera, language instruction, and proprioceptive hand state.

The entire observation space for the hand-level VLA policy is formalized as:
\begin{equation}
    o_t^{\text{hand}} = \left[ I_t^{\text{hand}},\ l_t,\ q_t^{\text{hand}},\ z_t^{\text{tac-f}},\ z_t^{\text{tac-s}} \right],
\end{equation}
where \( I_t^{\text{hand}} \) is the in-hand camera image, \( l_t \) is the language command, \( q_t^{\text{hand}} \) represents the hand joint state, and \( z_t^{\text{tac-f}}, z_t^{\text{tac-s}} \) are tactile embeddings capturing force magnitude and spatial contact patterns, respectively.

This multi-modal observation is passed to the action expert model, which is trained to predict hand action sequences \( A_t^{\text{hand}} \). The resulting policy is expressed as: 
\begin{equation}
    \pi_{\text{hand}}(A_t^{\text{hand}} \mid o_t^{\text{hand}}),
\end{equation}
which leverages both tactile sensing and vision-language context to achieve robust, force-adaptive grasping across a wide range of objects and conditions. The inclusion of complementary tactile cues enables the policy to modulate the grip force in response to contact events and object properties, significantly improving the grasping stability and success rate.


\subsection{Shared Autonomy for Data Collection}
Building upon DexGrasp-VLA, we introduce a shared autonomy framework for efficient data collection, combining the autonomous grasping capability of DexGrasp-VLA with the global guidance from the human for the pre-grasp interactions. The framework divides the shared control strategically: a human operator teleoperates the robotic arm’s end-effector via a VR interface for the obstacle-avoidant reaching and positioning, while the pre-trained DexGrasp-VLA policy autonomously controls the dexterous hand for fine grasping. This approach effectively reduces the operator’s cognitive load by eliminating the need to simultaneously coordinate both the arm and 12-DoF multi-finger hand, while preserving the naturalness and quality of the demonstrations. The reduction of the operator's cognitive load will be more substantial when it comes to bimanual operation with the same arm-hand setting.

\subsubsection{\textbf{VR-Based Arm Teleoperation System}}

\begin{figure}[t]
    \centering
    \vspace{-5mm}
    \includegraphics[width=1.0\textwidth]{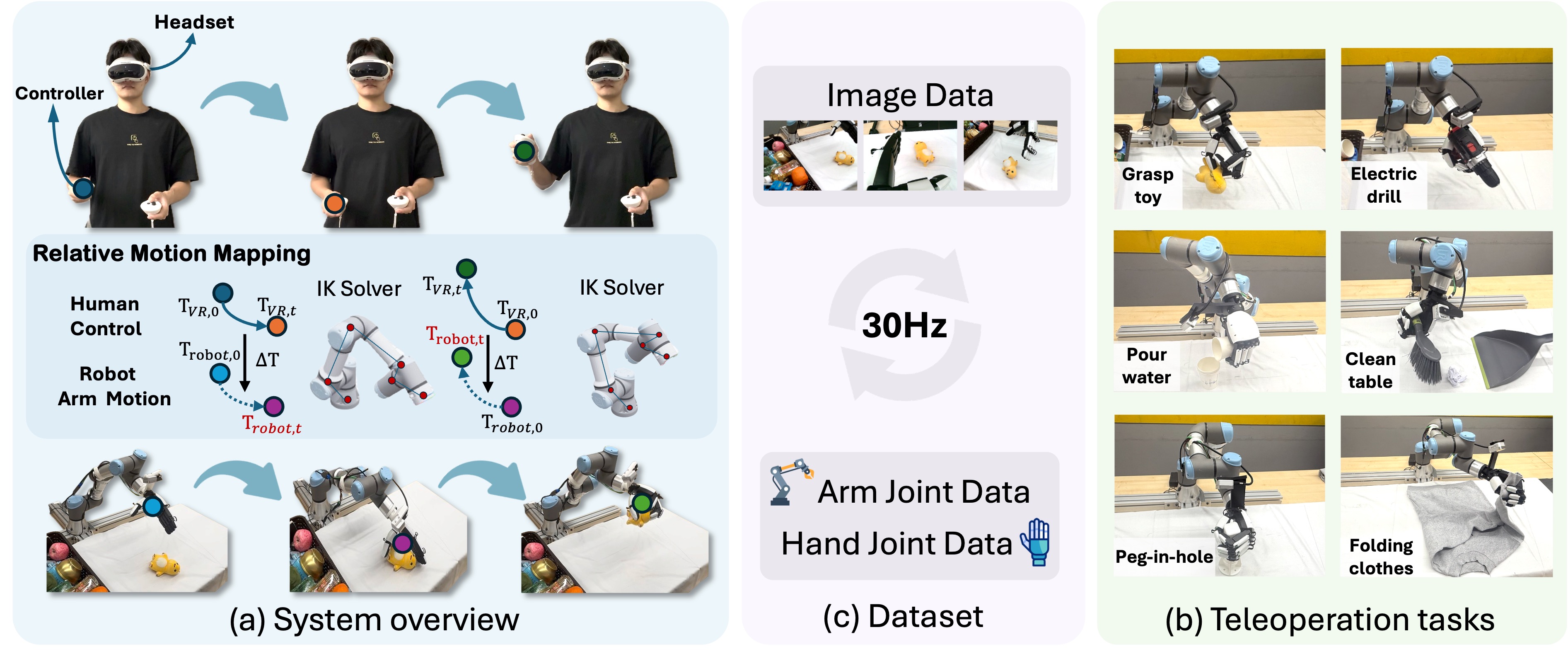}
    \caption{\normalsize Teleoperation system and task suits. A control approach based on relative motion mapping, which enables flexible teleoperation of diverse tasks and efficient data collection.}
    \label{fig:teleoperation}
\end{figure}


We implement a control paradigm based on relative motion mapping, built upon the XRoboToolkit [51] framework. This system is capable of flexibly handling diverse robotic tasks across short- and long-horizon operations, facilitated by a high-performance teleoperation framework with OpenXR support (90fps, <100ms latency) that enables real-time control and high-frequency data collection (at 30Hz), as shown in Fig.~\ref{fig:teleoperation}. Specifically, the tracking of the robot arm's end-effector is initiated through a ``clutch mechanism'' triggered by the user holding the controller's grip button. Upon activation, the initial poses of both the VR's controller, $T_{\text{VR},0} \in \text{SE}(3)$, and the robot arm's end-effector, 
$T_{\text{robot},0} \in \text{SE}(3)$, are recorded at the same time. Then, the change of poses, i.e., delta transformation ($\Delta T$) of the VR's controller from its initial pose $T_{\text{VR},0}$ to its current pose $T_{\text{VR},t}$, is continuously measured. This transformation is then composed with the initial pose of the robot arm (i.e., the end-effector pose at the moment when teleoperation is engaged) to calculate the new target pose, $T_{\text{robot},t}$. This relationship is formulated as:
\begin{equation}
T_{\text{robot},t} = T_{\text{robot},0} \cdot (T_{\text{VR},0}^{-1} \cdot T_{\text{VR},t}).
\end{equation}
An Inverse Kinematics (IK) solver is used to compute the target joint angles from this target pose, ensuring that the end-effector motion consistently maps the VR controller's displacement. Specifically, we formulate the inverse kinematics as a velocity-level Quadratic Program (QP). This optimization problem is solved using the PlaCo~\cite{placoweb} library, which is built on the Pinocchio~\cite{carpentier2019pinocchio} rigid body dynamics framework. The QP is formulated using the joint velocities $\dot{q}$ as optimization variables to find the optimal solution that minimizes the weighted sum of squared task errors, subject to a set of linear constraints:
\begin{gather} 
\min_{\dot{q}} \sum_{i=1}^{N} w_i \left\| J_i(q)\dot{q} + e_i(q) \right\|^2, \hspace{2mm} \text{s.t.} \quad l \le C(q)\dot{q} \le u \notag .
\end{gather}
Here, $q$ denotes the manipulator's joint configuration. Each task $i$ is characterized by its residual function $e_i(q)$, task Jacobian $J_i(q)$, and a corresponding scalar weight $w_i$. The matrix $C(q)$ represents additional hard constraints imposed on the system. Finally, the joint configuration $q$ is obtained through numerical integration. At each control time step $t$, the above QP is solved to compute the optimal joint velocity $\dot{q}$. This optimal velocity is then used to update the new joint angles for the next time step:
\begin{equation}
q_{t+1} = q_t + \dot{q} * \Delta t,
\end{equation}
where $\Delta t$ is the sampling time, and $q_t$ are the previous joint angles. This process iterates until the task errors $\|e_i(q)\|$ are below a threshold or a terminal condition is met. The VR-based teleoperation system captures and maps the motions of the VR controller in real time to the robotic arm's motion. This setup enables natural and responsive arm control in 3D space, significantly improving demonstration efficiency and precision.

\subsubsection{\textbf{Coordinated Arm-Hand Data Collection}}

For data collection, we engineered a multi-thread control architecture that enables seamless integration of human teleoperation (90 Hz) with autonomous policy execution (30 Hz), as well as streaming and saving time-synchronized data (30 Hz). In this setup, the human operator controls the arm's 6-DoF end-effector motion through a VR interface while the DexGrasp-VLA policy runs concurrently, generating appropriate grasping actions based on real-time visual and tactile feedback. This parallel execution mechanism preserves natural temporal coordination between arm's reaching and hand's grasping phases, enabling efficient collection of synegized arm-hand demonstrations. The collected datasets contain temporally synchronized observations and actions from both control sources, formally represented as:
\begin{equation}
\begin{gathered}
\mathcal{D}_\text{uni} = \{(o_t^\text{uni}, a_t^{\text{arm}}, a_t^{\text{hand}})\}_{t=1}^T; \hspace{2mm} a_t^{\text{arm}} \sim p_{\text{teleop}}, \quad
a_t^{\text{hand}} \sim \pi_{\text{hand}}(\cdot \mid o_t^{\text{hand}}),
\end{gathered}
\end{equation}
where the combined observation vector $o_t^\text{uni} = [I_t, l_t, q_t^{\text{arm}}, q_t^{\text{hand}}]$ incorporates multi-view RGB images $I_t$ (including static scene cameras and the eye-in-hand camera), language instruction $l_t$, and the joint states of both the arm $q_t^{\text{arm}}$ and hand $q_t^{\text{hand}}$.

This dataset structure provides several key advantages: (i) it maintains the natural coordination synergies between arm and hand movements that are characterized within human demonstration; (ii) it enables the learning algorithm to discover implicit coordination strategies from the demonstrated trajectories; and (iii) the rich multimodal observations support learning robust visuomotor policies that can generalize across object properties and environmental conditions. The dataset $\bm{ \mathcal{D}_\text{uni} }$ serves as the foundation for training end-to-end arm-hand manipulation policies that effectively combine human guidance for coarse macro-motions and the autonomous grasping for fine micro-movements of multiple fingers.


\subsection{Learning End-to-End Arm-Hand VLA Policy $\pi_{\text{uni}}$}

\begin{wrapfigure}{r}{0.55\linewidth}
    \centering
    \includegraphics[width=1.0\linewidth]{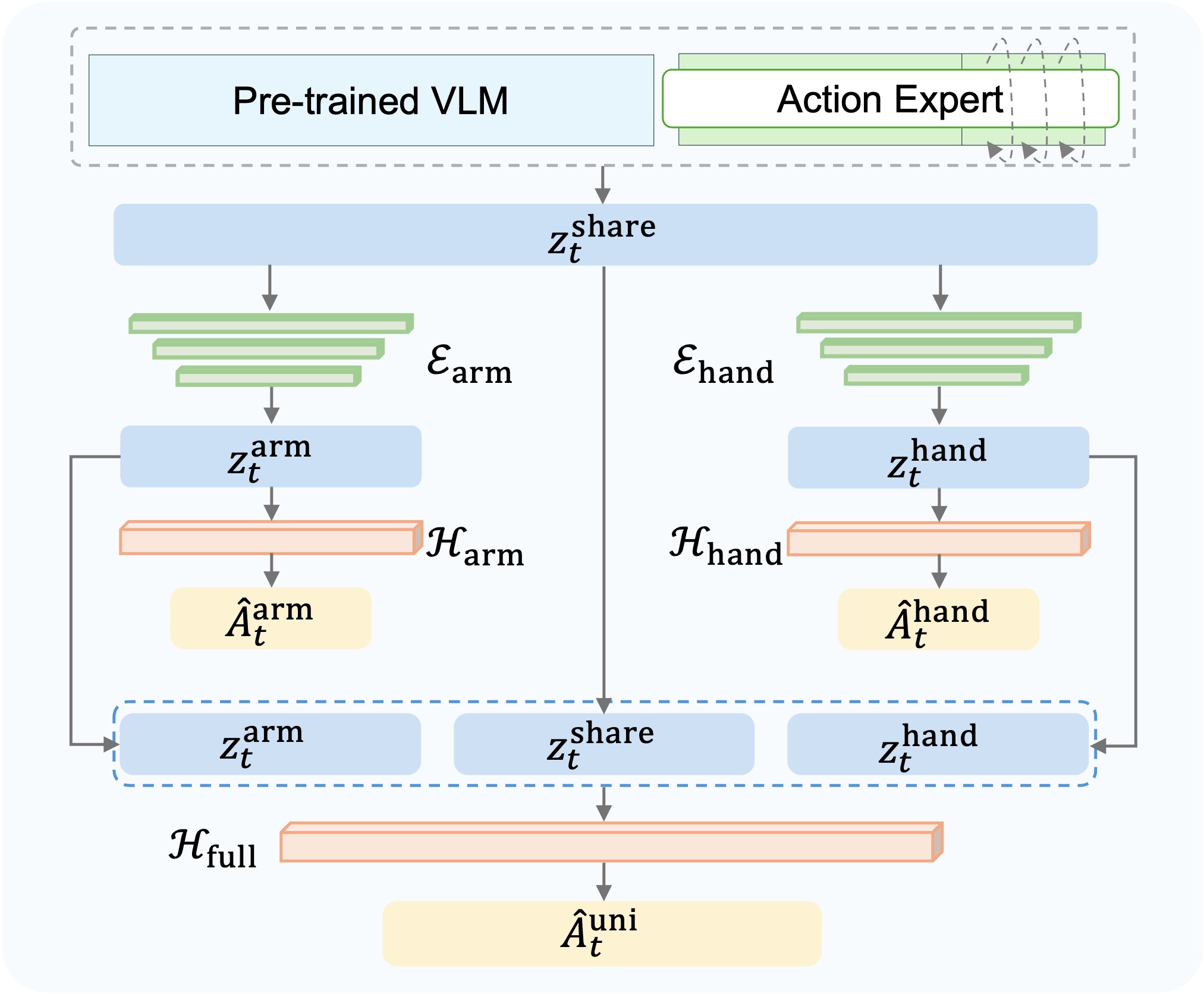}
    \vspace{-4mm}
    \caption{\normalsize Arm-hand feature enhancement for the VLA policy.}
    \label{fig:decoupled_framework}
    \vspace{-5mm}
\end{wrapfigure}

Building upon the arm-hand demonstration data collected through shared autonomy, we perform SFT of a pre-trained VLA model to learn an arm-hand coordinated dexterous grasping policy $\pi_{\text{uni}}(\mathcal{A}_{t}^\text{uni}\mid o_t^{\text{uni}})$. A key challenge lies in effectively handling the different and complementary characteristics of arm and hand motion. Our prior research efforts found that a vanilla version of SFT would only lead to limited performance, primarily because it fails to explicitly disentangle and represent the unique dynamics of the arm and hand, while also capturing their shared task context. 

To overcome this limitation, we developed the technical improvement that has no structural alteration as for the fine-tuning stage -- the Arm-Hand Feature Enhancement module. It extends the base architecture with explicit mechanisms to capture both shared contextual information relevant to coordinated arm-hand tasks, and distinct individual features that characterize arm and hand dynamics, respectively. Hence, it enables more effective and robust arm-hand coordination during dexterous grasping, for example, even under visual occlusion of one camera.

\subsubsection{\textbf{Arm-Hand Feature Enhancement}}

As illustrated in Fig.~\ref{fig:decoupled_framework}, our architecture builds upon the base $\pi_0$ model, which employs \texttt{PaliGemma} and \texttt{Gemma Expert} to encode multi-modal observations (including visual inputs, language instructions, and proprioceptive states) into a shared task representation $\bm{z_t^{\text{share}}} \in \mathbb{R}^{d_s}$. While this shared representation captures global task context, its native architecture is not built for controlling high-DoF dexterous hands and thus does not account for the distinct kinematics and dynamics of arm and hand movements. 

Reasoning from the fundamental differences in kinematic and dynamic interactions and functional roles, we developed the \textbf{Arm-Hand Feature Enhancement} module to resolve this issue, which explicitly models the complementary roles of macro-movement (arm) and micro-manipulation (hand). The shared representation $\bm{z_t^{\text{share}}}$ is passed through two dedicated multi-layer perceptrons (MLPs): $\bm{E_{\text{arm}}}$ and $\bm{E_{\text{hand}}}$. These produce limb-specific latent features $\bm{z_t^{\text{arm}}} \in \mathbb{R}^{d_a}$ and $\bm{z_t^{\text{hand}}} \in \mathbb{R}^{d_h}$, respectively. 

To ensure that these features capture meaningful limb-specific characteristics, we employ auxiliary prediction heads $H_{\text{arm}}$ and $H_{\text{hand}}$, which are trained to predict sub-actions $\hat{\mathcal{A}}_t^{\text{arm}}$ and $\hat{\mathcal{A}}_t^{\text{hand}}$ directly from their corresponding features. This auxiliary supervised learning through encoding-decoding ensures the features capture relevant underlying dynamics of their respective limbs (arm or hand), such as long-horizon reaching trajectories for the arm, and contact-rich and compliant motions for the hand.
For final action prediction, the main action head $H_{\text{main}}$ takes as input the fused representation
\(
\bm{z_t^{\text{fused}} = [z_t^{\text{share}}, z_t^{\text{arm}}, z_t^{\text{hand}}]},
\)
and outputs the unified action
\(
\hat{\mathcal{A}}_t^{\text{uni}} = [\hat{\mathcal{A}}_t^{\text{arm}}, \hat{\mathcal{A}}_t^{\text{hand}}].
\)
The use of $z_t^{\text{share}}$ preserves the global task context, while limb-specific features enable fine-grained adaptive control. This design allows the policy to balance the overarching task strategy with limb-level precision, resulting in coordinated and naturalistic motion.

\subsubsection{\textbf{Learning Objective}}

The model is optimized using a composite loss function that combines a main coordination loss with two auxiliary losses tailored to arm and hand specificity.

\paragraph{\textbf{Main Loss}} 
The primary objective is derived from the conditional flow matching, as used in $\pi_0$, and is applied to the full action sequence $\mathcal{A}_t^{\text{uni}} = (\mathcal{A}_t^{\text{arm}}, \mathcal{A}_t^{\text{hand}})$:
\begin{equation}
\mathcal{L}_{\text{main}}^\tau(\theta) =
\mathbb{E} \left[ \left\| \mathcal{H}_{\text{main}}(z_t^{\text{fused}}) - u(\mathcal{A}_t^{\tau,\text{uni}} \mid \mathcal{A}_t^{\text{uni}}) \right\|^2 \right],
\end{equation}
where $\mathcal{A}_t^\tau = \tau \mathcal{A}_t + (1-\tau)\epsilon$ is a noise-corrupted action chunk, $u(\mathcal{A}_t^\tau \mid \mathcal{A}_t) = \epsilon - \mathcal{A}_t$ is the target vector field.

\paragraph{\textbf{Auxiliary Expert Losses}}  

Two auxiliary losses are used to reinforce the disentanglement and specialization of arm and hand:
\begin{equation}
\mathcal{L}_{\text{hand}}^\tau(\theta) = \mathbb{E} \left[ \left\| \mathcal{H}_{\text{hand}}(z_t^{\text{hand}}) - u_{\text{hand}}(\mathcal{A}_t^{\tau,\text{hand}} \mid \mathcal{A}_t^{\text{hand}}) \right\|^2 \right],
\end{equation}
\begin{equation}
\mathcal{L}_{\text{arm}}^\tau(\theta) = \mathbb{E} \left[ \left\| \mathcal{H}_{\text{arm}}(z_t^{\text{arm}}) - u_{\text{arm}}(\mathcal{A}_t^{\tau,\text{arm}} \mid \mathcal{A}_t^{\text{arm}}) \right\|^2 \right].
\end{equation}
These losses ensure that the dedicated features $z_t^{\text{arm}}$ and $z_t^{\text{hand}}$ can predict their corresponding limb actions.

\paragraph{\textbf{Total Loss}}  

The overall training objective is a weighted sum:
\begin{equation}
\mathcal{L}_{\text{total}} = \mathcal{L}_{\text{main}} + \lambda \left( \mathcal{L}_{\text{hand}} + \mathcal{L}_{\text{arm}} \right),
\end{equation}
where $\lambda$ is a hyperparameter that balances between global coordination and limb-specific specialization. This objective encourages the model to learn a representation that is both integrative and limb-aware, leading to more robust and generalizable arm-hand coordination.

\subsection{Corrective Human-in-the-Loop Teleoperation System} \label{ssec:corrective-teleop}

\begin{figure}[t]
    \centering
    \includegraphics[width=1.0\textwidth]{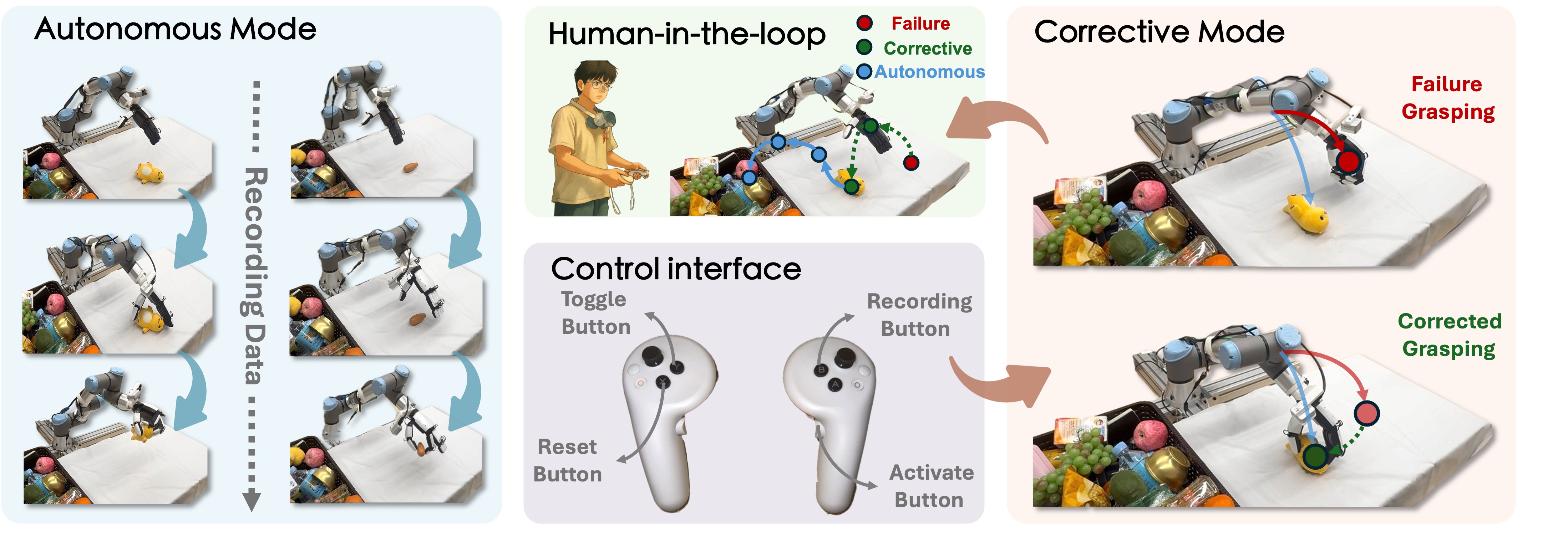}
    \vspace{-3mm}
    \caption{\normalsize Corrective human-in-the-loop teleoperation system enables efficient and effective human intervention to collect corner case data during the deployment phase, iteratively improving the policy robustness.}
    \label{fig:intervention}
    \vspace{-3mm}
\end{figure}

To enable robust deployment in unstructured environments where distribution shifts and long-tail scenarios are inevitable, as shown in Fig.~\ref{fig:intervention}, we implement a corrective human-in-the-loop teleoperation system with an incremental supervised fine-tuning (SFT) framework. This system enables continuous policy adaptation through real-world interaction and targeted data collection.

During deployment, the unified arm-hand policy $\pi_{\text{uni}}$ operates autonomously. Successful trajectories are recorded as positive demonstrations. When failures occur, the system toggles a shared-autonomy mode where a human operator intervenes via teleoperation to recover from failures. Both the failure episodes and subsequent recovery trajectories are curated as corrective demonstrations.

Formally, at each iteration $k$, the policy $\pi_{\text{uni}}^{(k)}$ collects a dataset $\mathcal{D}^{(k)} = \mathcal{D}_{\text{success}}^{(k)} \cup \mathcal{D}_{\text{corrective}}^{(k)}$, where $\mathcal{D}_{\text{success}}^{(k)}$ contains the data from successful autonomous policy rollouts, and $\mathcal{D}_{\text{corrective}}^{(k)}$ contains the data from human-guided failure recoveries. 

Equation \ref{eq:dataset-composition} delineates the composition of the training dataset, where corrective examples include both the failure context and the human-provided recovery strategy, enabling the policy to learn error detection and correction strategies:
\begin{equation} \label{eq:dataset-composition}
\mathcal{D}^{(k)} = \underbrace{\left\{ \left( o_t, a_t \right) \right\}}_{\mathcal{D}_{\text{success}}^{(k)}} \cup \underbrace{\left\{ \left( o_t^{\text{(fail)}}, a_t^{\text{(fail)}}, o_t^{\text{(rec)}}, a_t^{\text{(rec)}} \right) \right\}}_{\mathcal{D}_{\text{corrective}}^{(k)}}.
\end{equation}

We use the brief notation below (instead of the argmin form) to notate the policy update:
\begin{equation} \label{eq:iterative-update}
\pi_{\text{uni}}^{(k+1)} = \text{SFT}\left(\pi_{0} ; \mathcal{D}_{\text{uni}} \cup \mathcal{D}^{(k)} \right).
\end{equation}

This iterative process creates a self-improving cycle, where the policy progressively learns to handle increasingly challenging scenarios with an augmented dataset that has better coverage. The corrective demonstrations provide crucial learning signals for failure recovery, while successful trajectories reinforce robust behaviors. It shall be noted that the new model still directly derives from the original foundation model $\pi_{0}$, while the dataset has been constantly curated as $\mathcal{D}_{\text{uni}} \cup \mathcal{D}^{(k)} $, where $\mathcal{D}^{(k)}$ is updated via Equation \ref{eq:dataset-composition}.

The key advantage of this approach is its ability to efficiently address the long-tail problem in robotic manipulation. By focusing human intervention only on failure cases, it maximizes data collection efficiency while ensuring the policy's coverage on downstream tasks.

\section{Results}
\label{sec:results}
\subsection{Experimental Settings}
\subsubsection{\textbf{Manipulation Platform}} 

 \begin{figure*}[!h]
     \centering
     \vspace{-5mm}
     \includegraphics[width=\textwidth]{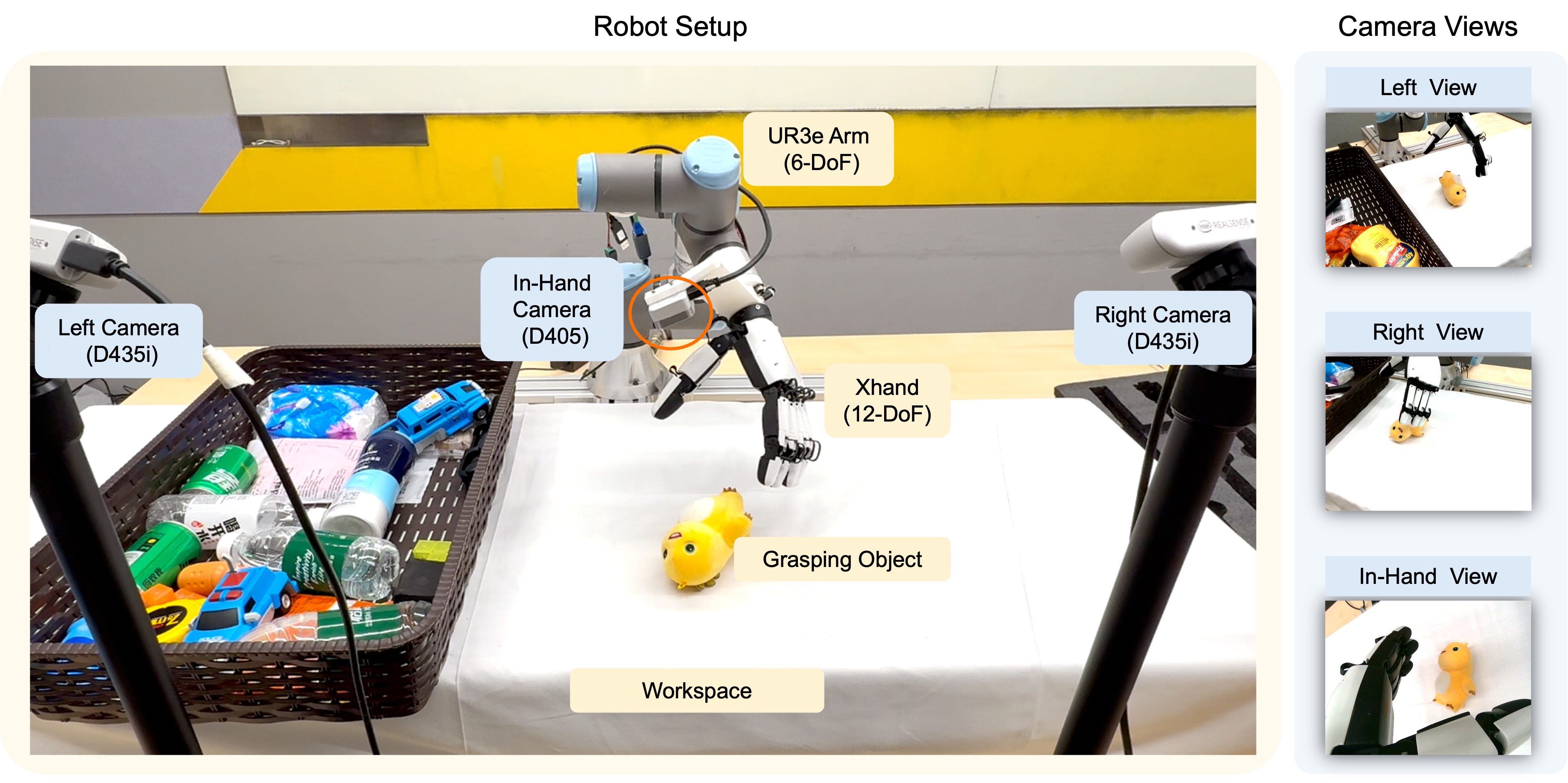}
     \vspace{-3mm}
     \caption{Hardware setup of the integrated robotic system. (1) Dexterous Manipulator: A 6-DoF UR3e arm with a five-finger 12-DoF \texttt{Xhand}, featuring fingertip tactile sensing for compliant manipulation. (2) Perception Suite: Three RGB-D cameras (two static, one wrist-mounted) for global scene perception and close-range visual observation.}
     \label{fig:setup}
     \vspace{-3mm}
 \end{figure*}
 
Our experimental platform, consisting of an integrated robotic manipulator with a hand and a multi-sensor perception system (as shown in Fig.~\ref{fig:setup}), provides the necessary physical and sensory capabilities to develop and evaluate our shared autonomy and learning algorithms for dexterous manipulation.

\textbf{Robot System:} The robot system consists of a UR3e collaborative robotic arm (6-DoF) mounted with a five-finger \texttt{Xhand} (12-DoF)~\cite{xhand}. Each fingertip of the \texttt{Xhand} incorporates an array of 120 tri-axial force sensors, providing high-resolution 3D contact feedback essential for compliant grasps and reactive control.

\textbf{Perception System:} The perception system employs 3 RGB-D cameras: two fixed Intel RealSense D435i cameras for global scene observation (binocular vision) and one wrist-mounted Intel RealSense D405 camera for eye-in-hand visual feeds needed for close-range manipulation.

\subsubsection{\textbf{Datasets}}

 \begin{figure*}[!h]
     \centering
     \vspace{-3mm}
     \includegraphics[width=\textwidth]{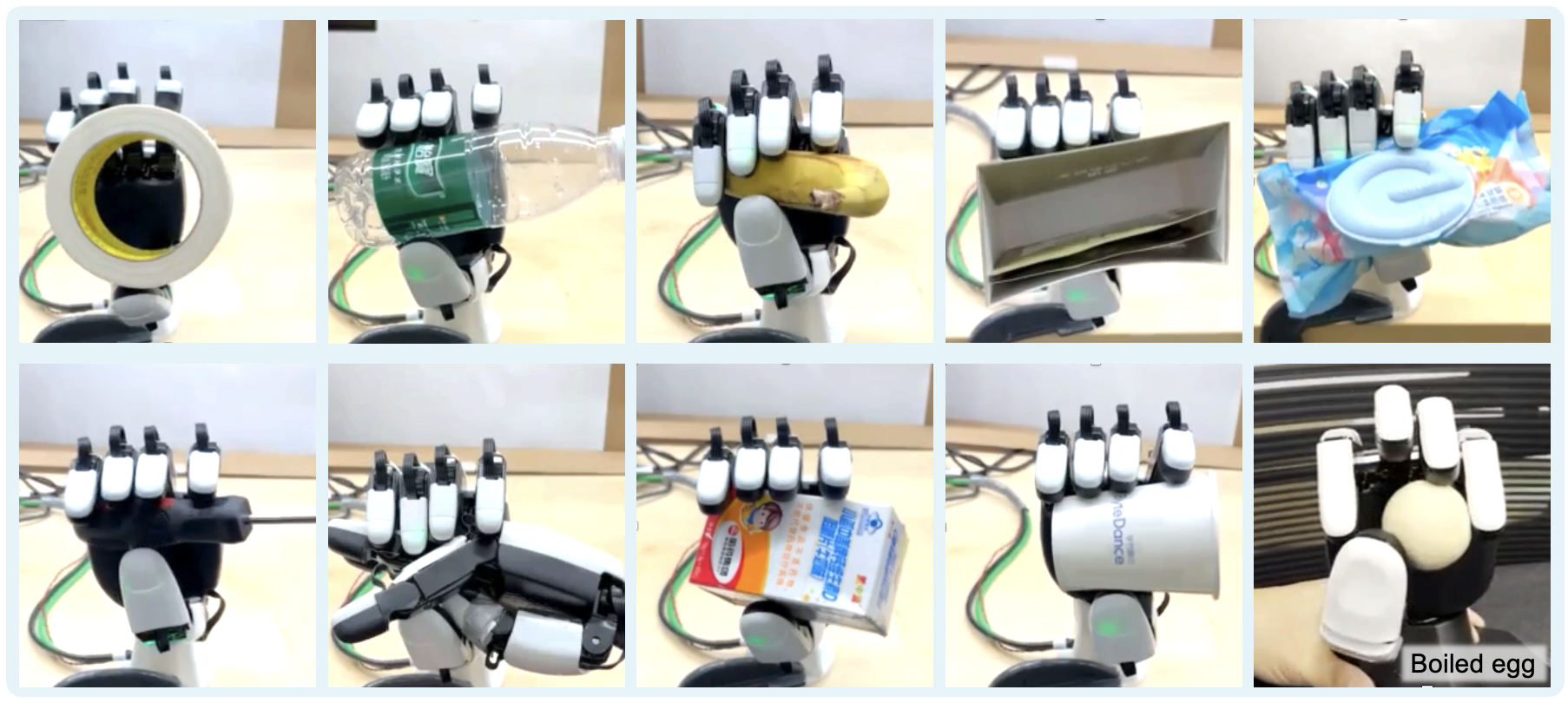}
     \caption{Force-adaptive grasping of different objects by the LSTM-based policy with tactile sensing. Using the tactile feedback, the LSTM policy learns admittance control by generating position control commands to regulate contact forces at fingertips, enabling firm grasps of very light delicate objects (paper cup, soft-boiled egg) as well as heavy rigid ones (a 1.2 kg of an unmounted robotic hand). }
     \label{fig:vis_lstm_grasp}
     \vspace{-0mm}
 \end{figure*}
 
We construct four distinct datasets to support the different training stages of our framework, each designed for a specific learning objective.

\textbf{LSTM Pretraining Dataset:} For training the initial blind grasping policy, we collect a hybrid dataset combining 150 human teleoperation demonstrations via Leap Motion retargeting (ie.e, retargeting from human hand to robot hand), and 68 autonomous trajectories generated by the force-adaptive controller (See Section \ref{ssec:lstm-policy} and Equation \ref{eq:force-control}). This dataset focuses on individual finger coordination and force modulation, containing time-series data of joint positions, torques, and tactile readings. As shown in Fig.~\ref{fig:vis_lstm_grasp}, the trained LSTM policy can ``blindly'' grasp various objects with force-adaptive capabilities using tactile sensing.

\textbf{DexGrasp-VLA Hand Policy Dataset ($\mathcal{D}_{\text{hand}}$):} We used handhold equipment to mount both the robot hand and the eye-in-hand camera on the handle, then let the human operator hold the robot hand and trigger the LSTM policy to grasp diverse objects in clutter. In this manner, a dataset consisting of 180 successful grasping trajectories is collected in cluttered scenes containing 60 different objects with diverse shapes and materials. Each trajectory streams synchronized eye-in-hand camera images, proprioceptive states, high-dimensional tactile data, and the corresponding action commands.

\textbf{End-to-end Arm-Hand VLA Dataset ($\mathcal{D}_{\text{uni}}$)}: As a pilot study, we collected 100 coordinated arm-hand demonstrations through our shared autonomy framework for end-to-end arm-hand policy training. These trajectories cover 20 common household objects in single-object scenes and include multi-view RGB observations, language instructions, and synchronized arm and hand action sequences.

\textbf{Corrective Intervention Datasets ($\mathcal{D}_{\text{orient}}$, $\mathcal{D}_{\text{corner}}$):} To support iterative policy refinement, we constructed two specialized datasets through human-in-the-loop corrective teleoperation. (1) $\mathcal{D}_{\text{orient}}$: 50 trajectories focused on recovering from orientation-specific grasping failures. (2) $\mathcal{D}_{\text{corner}}$: 50 trajectories addressing various challenging corner-case scenarios. These datasets are used to augment the primary training data and enhance policy robustness through targeted fine-tuning. 

All the aforementioned datasets include synchronized multi-modal observations (RGB images, proprioceptive states, and tactile readings) and action sequences at 30Hz, ensuring comprehensive coverage for learning complex manipulation behaviors across different training stages.

 \begin{figure*}[t]
     \centering
     \includegraphics[width=\textwidth]{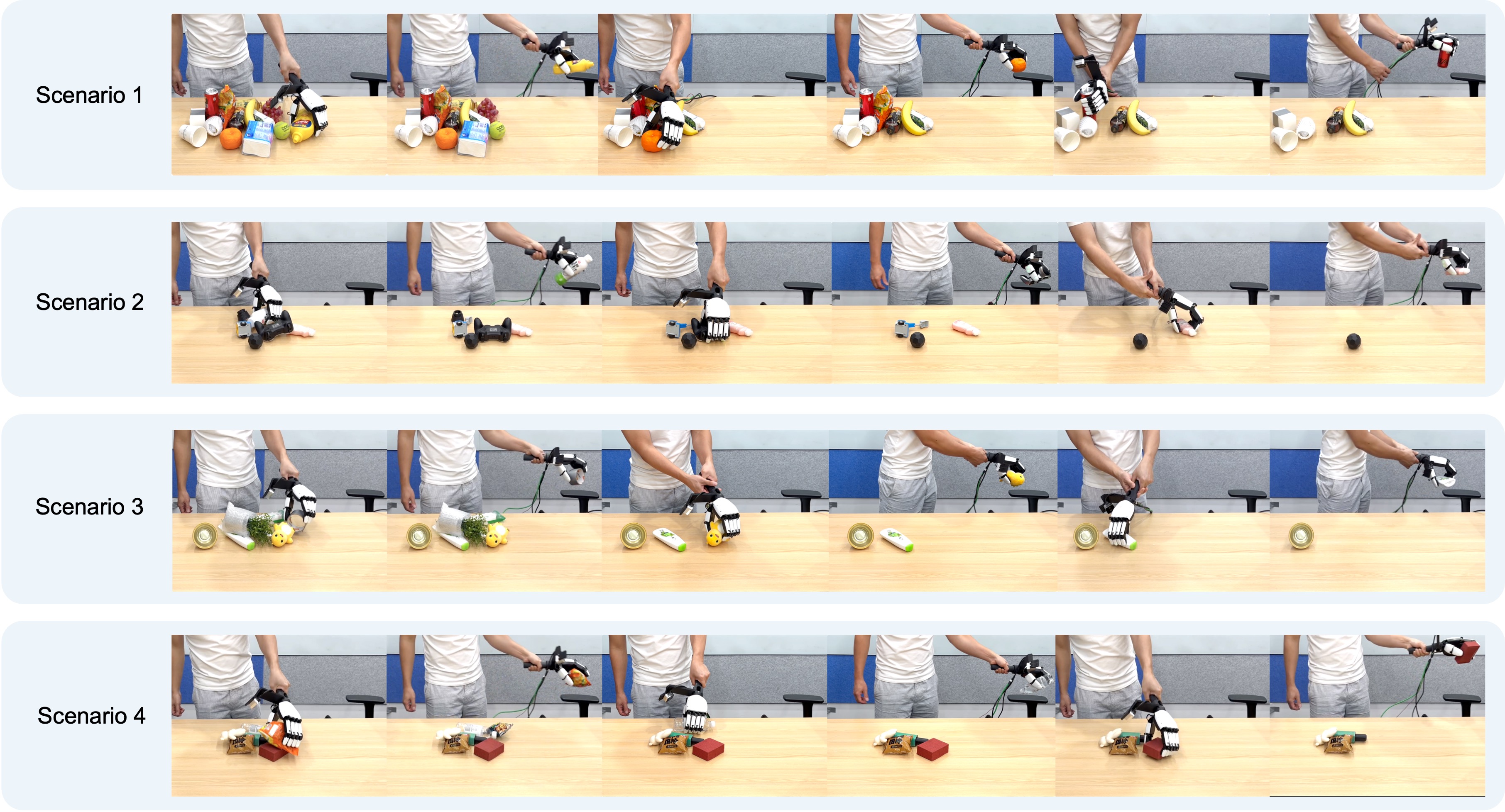}
     \caption{Demonstration of the DexGrasp-VLA policy for autonomous table bussing in clutter. The dexterous hand is operated in a handheld setup, and the DexGrasp-VLA policy performs autonomous and robust hand-level grasping based on its tactile and visual feedback. }
     \label{fig:vis_grasp}
     \vspace{-5mm}
 \end{figure*}

\subsection{Main Results}

\subsubsection{\textbf{Grasping Performance of the DexGrasp-VLA Policy \texorpdfstring{$\pi_{\text{hand}}$}{pi(hand)}}}

We evaluate the hand-only DexGrasp-VLA policy in cluttered tabletop scenarios, where the task requires picking and removing all the objects in clutter from the table. To facilitate testing, particularly for ablation studies on the hand policy itself, we adopt a \textit{handheld setup}: the human operator physically holds and moves the \texttt{Xhand} mounted on a handle to reach various objects, while the DexGrasp-VLA policy runs autonomously to decide when and how to execute grasping motions based on real-time visual, tactile, and proprioceptive inputs. This setup decouples the grasping capability of the hand from the errors and failures from the reaching control of the robotic arm, which allows for more specific evaluation on the finger-level DexGrasp-VLA policy.

Fig.~\ref{fig:vis_grasp} shows representative example trials. The evaluation covers five distinct scenes containing more than 50 objects, where a bucket of stocked objects was randomly dumped on the table, including both seen and unseen items with a large variation in size, color, shape, and material. The policy achieves a \textit{95.5\% overall success rate}, consistently clearing nearly all objects in each scene. These results confirm that the hand-level VLA policy robustly handles cluttered settings and generalizes across diverse object properties. Using its autonomous performance and generalization ability, we integrated this policy as the \textit{VLA Copilot} within our \textit{Shared Autonomy} framework. This allows efficient and scalable collection of high-quality, generalizable grasping demonstrations with minimal human effort, by exploiting the policy's reliable grasping ability under real-world variability.

\subsubsection{\textbf{Grasping Performance of the VLA Policy \texorpdfstring{$\pi_{\text{uni}}$}{pi(uni)}}}

\begin{wraptable}{r}{0.5\linewidth} 
  \centering
  \vspace{-5mm} 
  \small
  \setlength{\tabcolsep}{3pt} 
  \caption{Success rates (\%) of the final end-to-end arm–hand VLA policy $\pi_{\text{uni-final}}$ across 50 objects.}
  \label{tab:vla_succ}
  \begin{tabular}{c|c|c|c}
    \toprule
    \textbf{Methods} & \textbf{Seen objects} & \textbf{Unseen objects} & \textbf{Average} \\
    \midrule
    $\pi_{\text{uni-final}}$ & 91.7 & 85.6 & 88.7 \\
    \bottomrule
  \end{tabular}
  \label{tab:final_vla}
\end{wraptable}

  \begin{figure*}[t]
     \centering
     \includegraphics[width=1.0\textwidth]{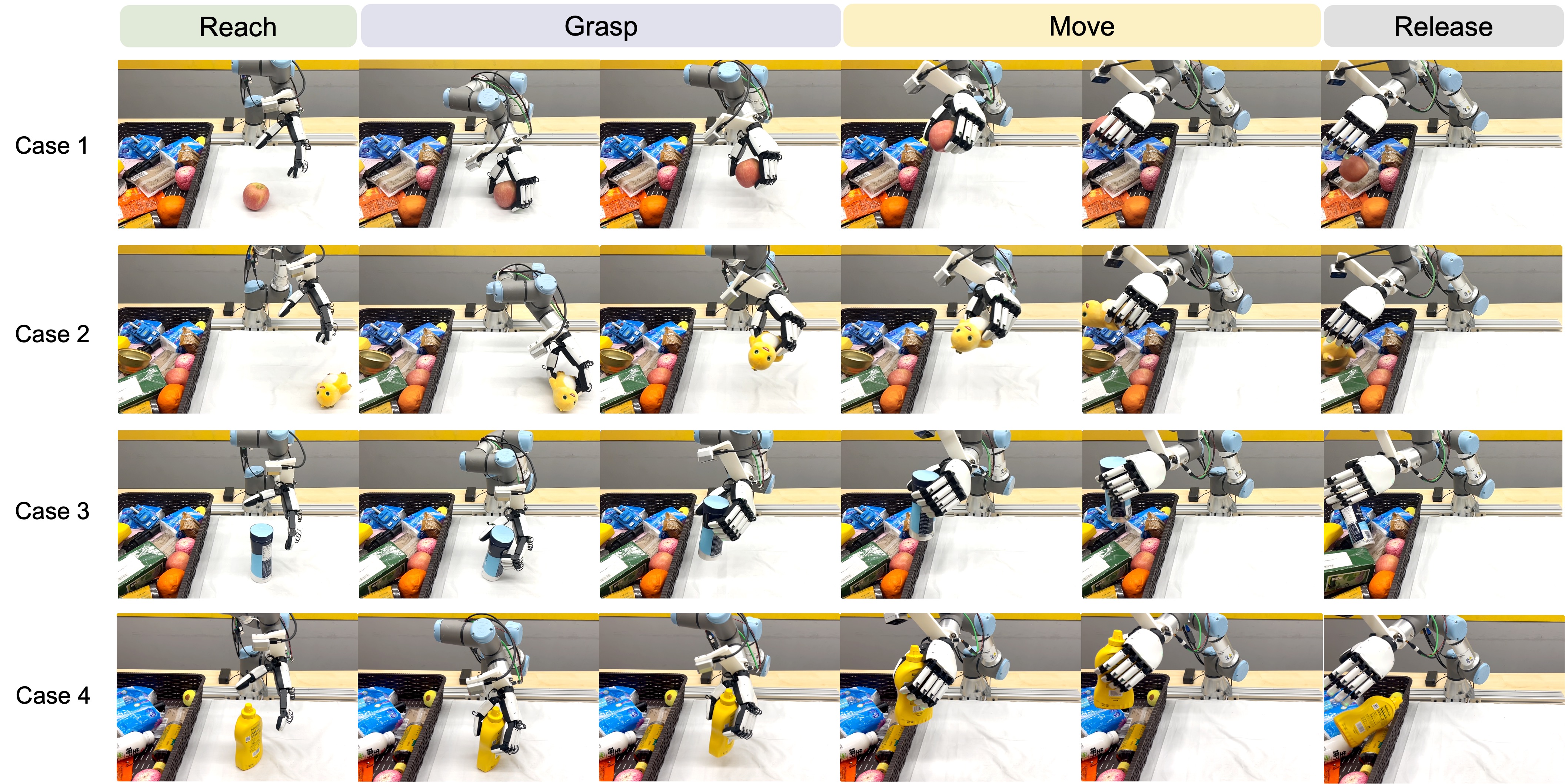}
     \caption{Grasping and placing various objects via end-to-end (arm-hand) VLA policies with a dexterous hand. (See \textbf{Appendix \ref{sec:app:addition}} for additional real robot tests.)}
     \label{fig:vis_pnp}
     \vspace{-4mm}
 \end{figure*}

\begin{figure}[!h]
    \centering
    \includegraphics[width=1.0\textwidth]{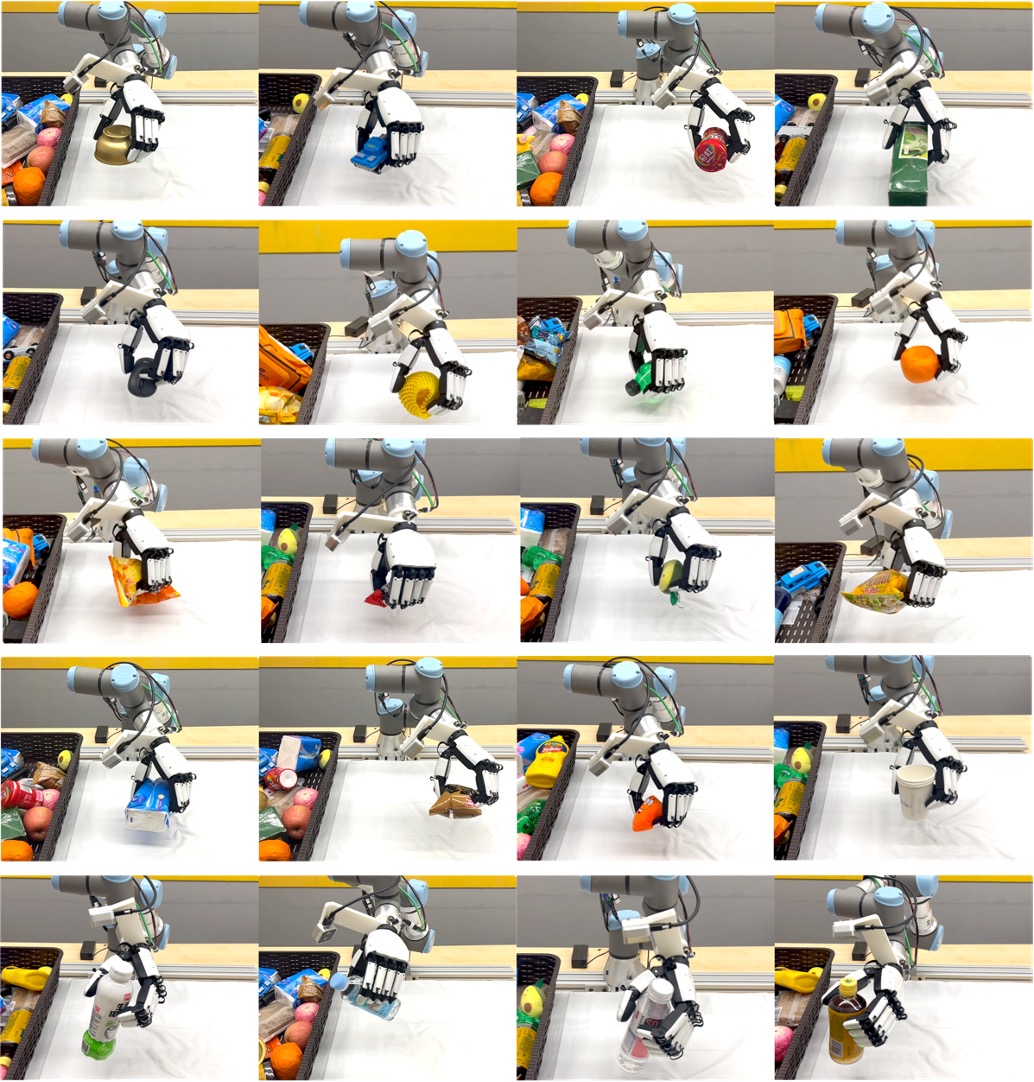}
    \caption{Grasping diverse objects with variations in size, color, and material properties.}
    \label{fig:diverse_objects}
    \vspace{-5mm}
\end{figure}

We evaluate the end-to-end arm-hand VLA policy, $\pi_{\text{uni}}$, on a pick-and-place task involving 20 seen objects and more than 30 unseen objects with diverse shapes, sizes, and materials. Each object is tested in 3 trials with randomized positions and orientations within a 40cm × 40cm workspace. In each trial, the robot attempts to grasp the object and place it in a target basket. A trial is deemed successful if the object is securely grasped and accurately placed without slipping or dropping. As summarized in Table~\ref{tab:final_vla}, $\pi_{\text{uni}}$ achieves an average success rate of 88.7\% across all 50 objects, with 91.7\% on seen objects and 85.6\% on unseen objects. More results of dexterous grasping can be found in \textbf{Appendix~\ref{sec:grasp_more}}. 

Fig.~\ref{fig:vis_pnp} demonstrates that the policy delivers consistent and reliable performance under significant object variations. The model demonstrates high performance consistently on familiar objects, reflecting effective hand-arm coordination and stable grasping capabilities. Furthermore, $\pi_{\text{uni}}$ generalizes robustly to novel objects and challenging orientations, enabling reliable pick-and-place across diverse object geometries and configurations (See more results across different robotic end-effector in logistics packing and industrial assembly in \textbf{Appendix~\ref{sec:app:addition}}).

It is important to note that these results were achieved with an initial proof-of-concept of the full stack, without particular engineering efforts. Therefore, it is realistic to anticipate that with more dedicated investments in optimizing the robotic hardware and its software, the system-level performance built on the VLA control architecture can reach the next level of near or above 95\% success rate, and thus be ready for real-world deployments, where the $>95\%$ success rate suffices and can be adopted in the industrial Standard Operating Procedure (SOP).

\subsection{Ablation study}
  \begin{figure*}[t]
     \centering
     \includegraphics[width=1.0\textwidth]{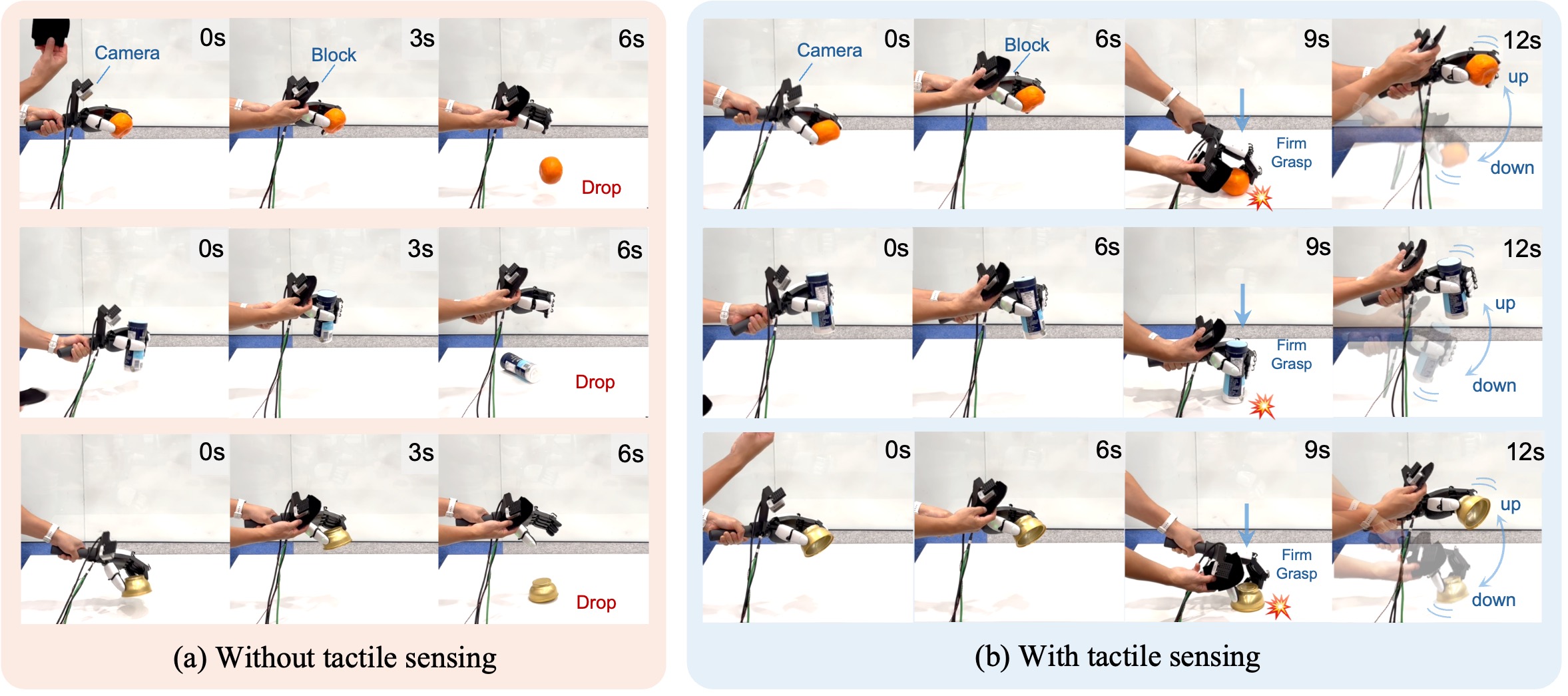}
     \caption{Effectiveness of tactile-based DexGrasp-VLA $\pi_{\text{hand}}$ and robust grasping performance with tactile sensing under camera occlusion: (a) Without the sense of touch (tactile), the object slips while camera is occluded; (b) With dual tactile features, the policy demonstrates a sustained, stable grasp throughout the test without any slippage, even under external perturbations (bumping against the table at 9s, and intense up-down shaking from 9s-12s).}
     \label{fig:vis_tac}
     \vspace{-0mm}
 \end{figure*}

\subsubsection{\textbf{Effectiveness of Tactile Sensing in the VLA Policy \texorpdfstring{$\pi_{\text{hand}}$}{pi(hand)}}}

To quantitatively evaluate the contribution of tactile sensing to grasp stability, we designed a two-phase test. The robot must first successfully grasp the object, then maintain a secure hold for 3 seconds with full vision, followed by 10 seconds under complete visual occlusion. A trial is considered successful only if the object is retained throughout both phases without being dropped. This protocol specifically evaluates the policy's dependency on tactile feedback when visual input is unavailable.

 \begin{table*}[!h]
  \centering
  \small
  \setlength{\tabcolsep}{3pt} 
  \caption{Success rates of dexterous grasping policy $\pi_{\text{hand}}$ across 10 everyday objects under different tactile configurations. }
  \begin{tabular}{c|c c c c c c c c c c c}
    \toprule
    Methods & Can & Bottle 1 & Bottle 2 & Apple & Orange & Banana & Bowl & Mug & Ball & Gamepad & Average \\
    \midrule
    $\pi_{\text{hand-origin}}$($\pi_0$~\cite{black2410pi0})    & 1/10  & 2/10 & 2/10 & 0/10  & 5/10 & 0/10 & 1/10 & 5/10 & 3/10 & 2/10 & 21\% \\
    $\pi_{\text{hand-tacf}}$       & 8/10  & 8/10 & 7/10 & 6/10  & 9/10 & 4/10 & 7/10 & 9/10 & 6/10 & 6/10 & 70\% \\
    $\pi_{\text{hand-tacf-tacs}}$  & 8/10  & 10/10 & 10/10 & 8/10  & 10/10 & 7/10 & 9/10 & 10/10 & 10/10 & 8/10 & 90\% \\
    \bottomrule
  \end{tabular}
  \vspace{-3mm}
  \label{tab:tac_vla_succ}
\end{table*}

As reported in Table~\ref{tab:tac_vla_succ}, tactile input is critical for robustness. Without any tactile feedback, the success rate drops to 21\%. Using only resultant force features ($f^{\text{tac-f}}$) improves performance to 70\%, and further incorporating \textit{both force and spatial tactile features} ($f^{\text{tac-f}}$ and $f^{\text{tac-s}}$) achieves 90\% success. This shows that tactile sensing is not only beneficial but also essential for sustaining stable grasps under perceptual uncertainty. The spatial tactile features enable the policy to detect and compensate for local contact slippage and shifts, while the force magnitude features are used to keep an appropriate level of grip forces.

\begin{figure*}[t]
    \centering
    \includegraphics[width=1.0\textwidth]{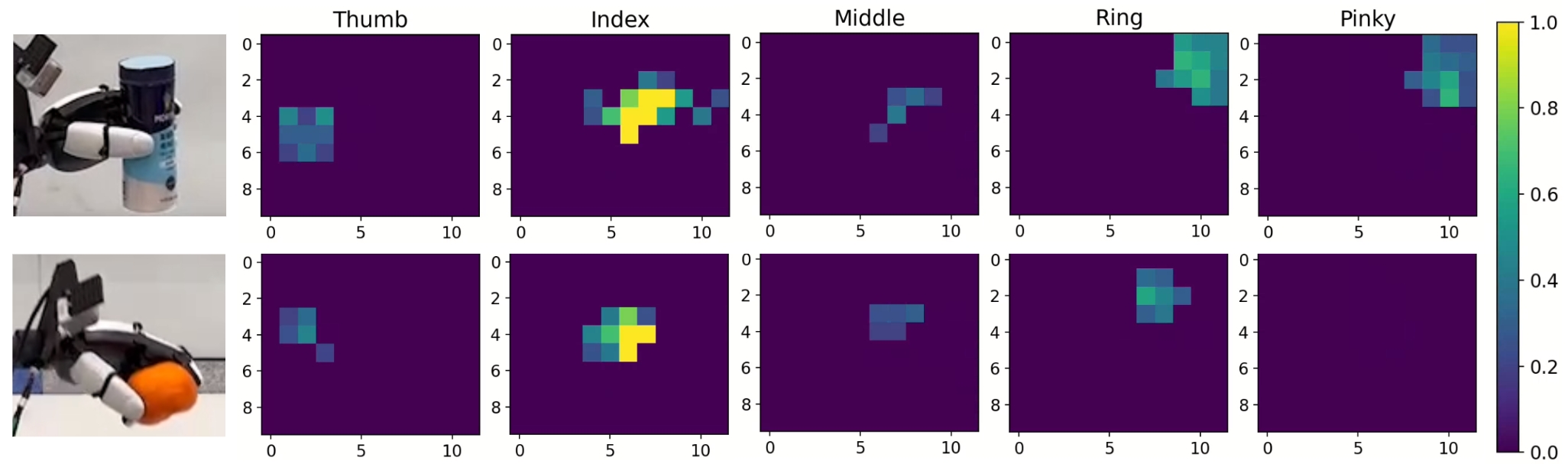}
    \caption{\normalsize Representative cases of grasping cylindrical and spherical objects, visualizing the distribution of surface contacts measured by tactile sensors at fingertips.}
    \label{fig:tactile}
    \vspace{-2mm}
\end{figure*}

Fig.~\ref{fig:vis_tac} illustrates a clear comparison: under visual occlusion, policies \textit{without} tactile feedback fail rapidly due to the lack of visuals, whereas the dual-tactile policy maintains a \textit{firm grip} under external perturbations (e.g., shaking or hitting the table). By visualizing tactile contacts (see the cases in Fig.~\ref{fig:tactile}) in real time, we can see that the adaptive adjusting of forces across the fingertips by the tactile-based policy. These results highlight that tactile feedback adds critical additional information to vision, enabling continuous modulation of grip forces and robust manipulation even when vision is degraded or occluded.

\begin{figure}[t]
    \centering
    \includegraphics[width=1.0\textwidth]{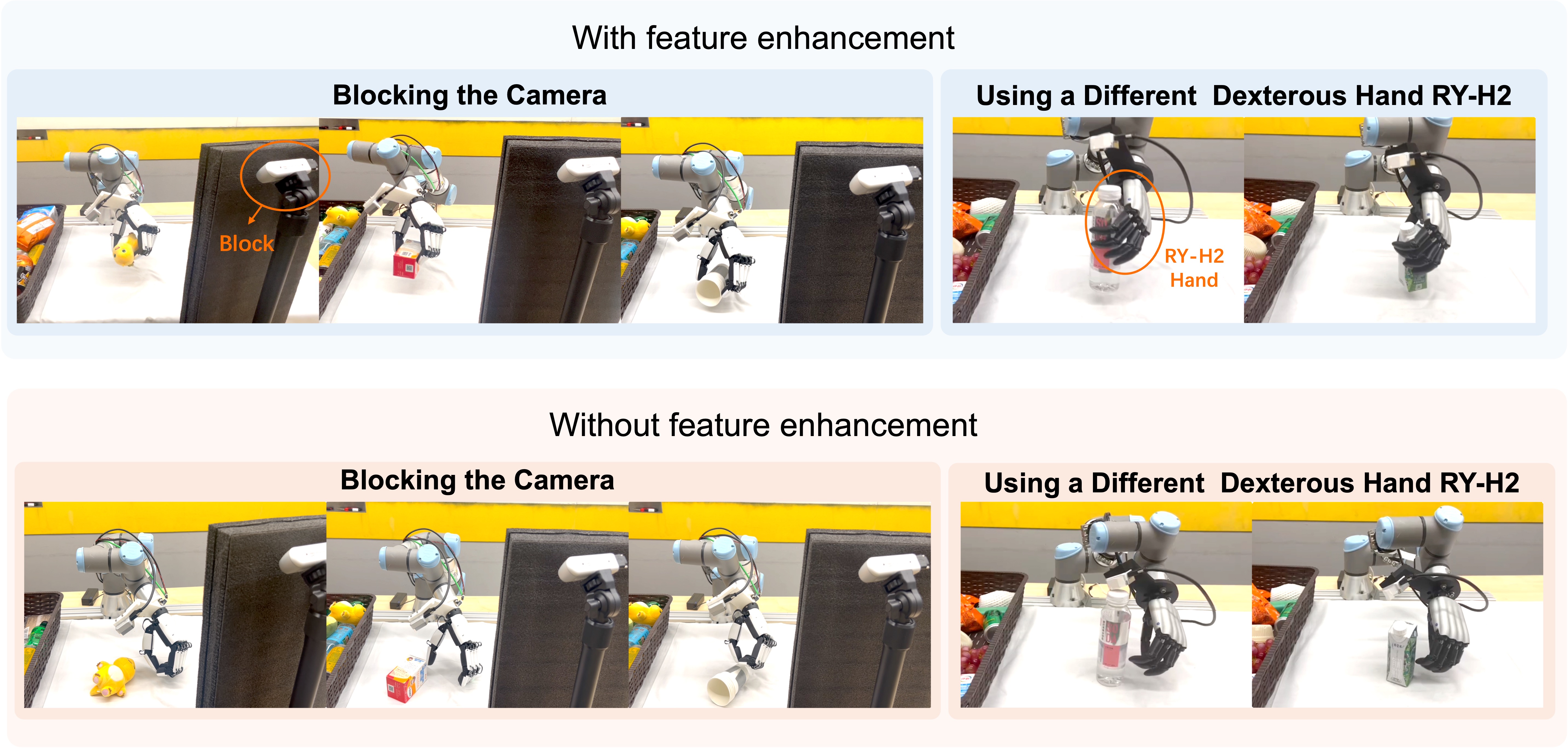}
    \caption{\normalsize Effectiveness of the Arm-Hand Feature Enhancement module. Under limited perception with a blocked camera, the enhanced policy enforces complementary roles: arm for reaching, and hand for grasping. This leads to more stable execution and a substantially higher grasp success rate than the baseline VLA. The approach generalizes across different dexterous hands, improving arm–hand coordination and grasp robustness on challenging objects.}
    \label{fig:de_vis}
    \vspace{0mm}
\end{figure}

\begin{figure}[!h]
    \centering
    \includegraphics[width=1.0\textwidth]{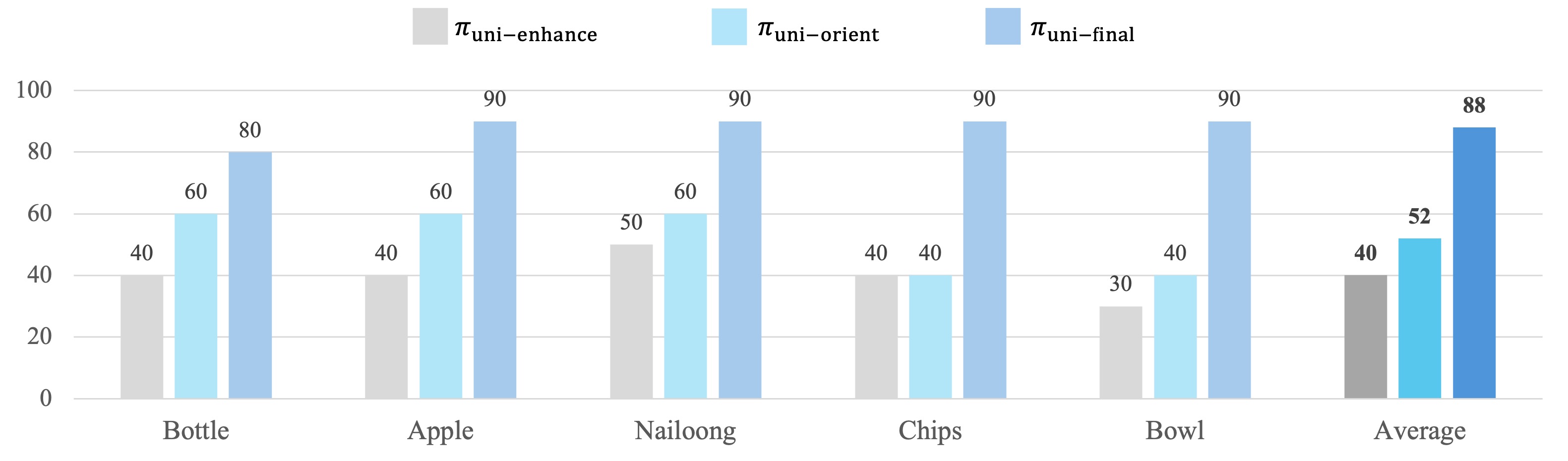}
    \caption{\normalsize Benchmark of iterative improvements enabled by corrective teleoperation. The initial policy $\pi_{\text{uni-enhance}}$ struggles with orientation and corner-case failures, the orientation-tuned policy $\pi_{\text{uni-orient}}$ improves generalization to varied poses, and the final policy $\pi_{\text{uni-final}}$ achieves robust success across all scenarios through targeted corrective demonstrations.}
    \label{fig:self_improvement_succ}
    \vspace{-5mm}
\end{figure}

\subsubsection{\textbf{Effectiveness of Arm-Hand Feature Enhancement in the VLA Policy \texorpdfstring{$\pi_{\text{uni}}$}{pi(uni)}}}

We conducted a comprehensive ablation study in 3 experimental settings, as shown in Table~\ref{tab:improve}, to evaluate the effectiveness of the proposed Arm-Hand Feature Enhancement module. The enhanced policy $\pi_{\text{uni-enhance}}$ is compared against the baseline $\pi_{\text{uni-origin}}$(initialized from $\pi_0$~\cite{black2410pi0}), with both models fine-tuned on the same dataset $\mathcal{D}_{\text{uni}}$. Results across three settings highlight consistent performance gains, and the comparison result is shown in Fig.~\ref{fig:de_vis}. On the main \texttt{Xhand} platform, $\pi_{\text{uni-enhance}}$ achieves a 95\% success rate compared to 88\% for the baseline, confirming improved coordination and grasp stability. When transferred to the \texttt{RY-H2} dexterous hand~\cite{ruiyan}, the enhanced policy outperforms the baseline by 10\% (81\% vs. 71\%), showing that the framework is applicable to different hardware that varies in kinematics and actuation (See more specifications of dexterous hands in \textbf{Appendix~\ref{sec:supplementary:hardware}}).

\begin{wraptable}{r}{0.5\linewidth}
  \centering
  \small
  \setlength{\tabcolsep}{3pt}
  \caption{Grasp success rates (\%) on 10 objects of the end-to-end arm–hand VLA policy $\pi_{\text{uni}}$.}
  \label{tab:pnp_succ}
  \begin{tabular}{c|c c c}
    \toprule
    \textbf{Methods} & \textbf{XHand} & \textbf{RY-H2} & \textbf{XHand-Occlude} \\
    \midrule
    $\pi_{\text{uni-origin}}\ (\pi_0~\cite{black2410pi0})$ & 88 & 71 & 19 \\
    $\pi_{\text{uni-enhance}}$                              & 95 & 81 & 58 \\
    \bottomrule
  \end{tabular}
  \label{tab:improve}
\end{wraptable}

To further assess the robustness to perceptual degradation, the right-view camera is \textit{occluded}, significantly impairing visual cues near contact regions. Under this challenging condition, compared to the baseline performance that drops drastically to 19\%, our $\pi_{\text{uni-enhance}}$ still maintains a 58\% success rate.

These consistent gains and advantageous robustness come from the explicit disentanglement of arm and hand representations. Using limb-specific feature pathways, the module prevents the fused representation from being dominated by global visual cues alone, which are especially vulnerable under occlusion. Instead, the arm branch encodes and inherits a large portion of its kinematics information, which remains reliable even with partial visual loss, while the hand branch captures fine-grained contact information. This complementary encoding allows the policy to preserve proprioception more specifically, thereby being less sensitive to vision degradation and maintaining stable grasp execution. 

In contrast, the baseline representation entangles all modalities into a single latent vector, making it harder to adaptively reweight information sources when one modality becomes unreliable. These results unveil that our proposed feature enhancement module not only improves coordination on the main \texttt{Xhand} platform, but also strengthens generalization across different hardware and enhances resilience to sensory uncertainty.

\subsubsection{\textbf{Effectiveness of Corrective Teleoperation}}
To evaluate the efficacy of our corrective human-in-the-loop teleoperation system, we designed a benchmark test with an expanded workspace (40 × 40 cm) containing objects arranged in a 3×3 grid with randomized orientations (upright and inverted). For a quantified evaluation, this setup has sufficient diversity in object placement and configuration, and presents significant challenges for policy generalization. As shown in Fig.~\ref{fig:self_improvement_succ}, the initial $\pi_{\text{uni-enhance}}$ model, trained only on $\mathcal{D}_{\text{uni}}$, failed on specific orientations and all corner-case scenarios. 

After fine-tuning with \textit{50 corrective trajectories} targeting orientation failures ($\mathcal{D}_{\text{orient}}$), the resulting $\pi_{\text{uni-orient}}$ model showed significantly improved handling of varied orientations and even demonstrated emergent error recovery behaviors. However, it remained ineffective for other corner cases, particularly those involving previously unseen locations that are out of distribution in the original training dataset. As a data-driven paradigm, a subsequent fine-tuning iteration with an additional 50 trajectories that specifically address these corner-case scenarios ($\mathcal{D}_{\text{corner}}$) produced a better version of a final model, $\pi_{\text{uni-final}}$, which successfully generalized across all test conditions.

Despite the small-scale benchmark of three policies across five objects in the easy-to-quantify setup, the results of success rates, e.g., from 20\% to 50\% differences, are very indicative to validate that our \textit{Corrective Teleoperation} system can identify and address specific failure modes through targeted data collection. More importantly, the results clearly show that iterative injection of human corrective demonstrations significantly enhances the robustness of the policy, enabling a form of \textit{continuous adaptation} to complex real-world scenarios that were not covered beforehand, or difficult to foresee and prepare in advance during the initial training data collection. 

To be truly useful and impactful in the real world, such a VLA-based learning framework must perform robustly on a variety of tasks and data collection approaches. To this end, we further validated its generality in both long-horizon sequential tasks and an industrial peg-in-hole assembly task. The latter, in particular, demands more spatial precision (i.e., $<$1-1.5 mm) beyond standard pick-and-place, involving complex contacts and fine motion sequences.

Through these studies, we confirmed that the corrective mechanism is effective not only with human teleoperation but also with alternative correction frameworks, such as \textit{fully automated motion planning}. In all cases, the framework can integrate targeted corrections from diverse data sources to efficiently improve policy performance. Additional experimental setups and results from the conventional robotic gripper and automated motion planning are provided in \textbf{Appendix~\ref{sec:appendix:gripper}}. Together, these findings show that our method offers a flexible and effective strategy for adapting and refining VLA-based policies in complex, real-world settings, showing the potential for broad applicability across task categories, end-effector morphologies (2-finger gripper and dexterous hands), corrective intervention, data augmentation and model re-training.

\section{Discussion, Future Work and Concluding Remarks}
\label{sec:conclusion}

\subsection{Discussion}
This work presents a VLA data collection and training framework via \textit{Shared Autonomy} which combines human teleoperation for global arm motion with an autonomous DexGrasp-VLA control of the hand that uses tactile feedback for adaptive grasping. This design reduces operator cognitive load and enables efficient collection of high-quality demonstrations, making it suitable for agile algorithmic developments. To resolve the challenge of fine action control of the multi-finger hand and enhance system robustness, we propose an \textit{Arm–Hand Feature Enhancement} module and a \textit{Corrective Human-in-the-Loop Teleoperation}, which altogether support fast policy iteration, policy re-training, and continual improvement. Experiments validated the effectiveness of this framework, achieving about 90\% grasping success rate in general among more than 50 seen and unseen objects, showcasing both efficiency and generalization.

In this work, we focus on grasping tasks as a fundamental capability in robotic manipulation. The high degree-of-freedom and complex multi-contact dynamics of five-finger dexterous hands impose a significant challenge, where numerous factors, from tactile sensing to arm-hand coordination, are deeply intertwined. Therefore, grasping serves as an ideal testbed, providing a well-defined yet challenging benchmark to validate shared autonomy, multimodal policy learning, and corrective teleoperation under diverse physical contact conditions. It allows for a methodological focus to systematically isolate and investigate the role of each core component, such as a more structured algorithmic integration of multimodal sensorimotor pathways for tactile sensing and different techniques for shared autonomy, without involving the extra complexity of dexterous in-hand manipulation. This focused approach enables a clear validation of our framework, establishing a foundational understanding of multimodal policy learning that paves the way for developing and scaling up to more sophisticated large-scale foundational models with human-level dexterity.


Despite these promising results as a zero-to-one proof of concept, limitations remain: the current system has not been tested in more complex long-horizon manipulation, such as tool use or in-hand reorientation; the full tactile integration, while critical for grasp stability, remains challenging due to noise and feature misalignment with arm movements (reaching motions without touching); and corrective mechanism and continual improvements still rely on human intervention, which limits the scalability and thus fully autonomous approaches are clearly more promising at a larger scale.

\subsection{Future Work}
The proposed architecture and methodology are designed with inherent extensibility, which allows a cohesive way for future upgrade with more advanced hardware, e.g., the VLA copilot system and the teleoperation are combinatorial. The shared autonomy interface naturally separates high-level human guidance from low-level execution, allowing new VLA copilots for various skills to be integrated within the same framework, for example, more sophisticated dexterous in-hand manipulation with multi-finger hands. Similarly, the policy structure, particularly the arm–hand feature enhancement, is task-agnostic and can adapt to diverse demonstrations, while the corrective teleoperation framework applies readily to new tasks.

Future work would call for a community effort to improve the VLA paradigm in multiple directions. This framework can be extended to a broader set of manipulation skills, including object re-orientation, insertion, and long-horizon tasks, by developing or combining additional specialized AI-driven copilots. We will also investigate advanced tactile representation learning techniques to better fuse tactile, visual, and proprioceptive inputs, for example, through selective routing or conditional gating, and explore auxiliary prediction losses inspired by the feature enhancement module.

Furthermore, future work can focus on autonomous error detection and composite recovery strategies. First, the human-in-the-loop teleoperation correction phase can be replaced by an autonomous error recovery module. Reinforcement learning and self-supervised adaptation can be used to train a model to autonomously detect and recover from errors, thus reducing the dependence on human intervention. Second, the coordination and switching of control authority between the human and the robot can be dynamically and seamlessly adjusted based on the perception of the context, such as task difficulty, environmental safety, and human state. This perceptual capability would be achieved via VLMs, and allow the system to autonomously determine when it can autonomously operate error recovery by itself, and when it needs human input or intervention.

\subsection{Conclusion}
In summary, this study serves as a practical handbook for developing an effective framework for dexterous manipulation learning, which is validated through extensive grasping experiments and is suitable for agile research development with minimal requirements of manpower. Such a framework demonstrates its effectiveness at a system-level of engineering and integration, using techniques ranging from force-adaptive compliance control, shared autonomy, human-in-the-loop teleoperation, VR-based human-robot interface, to multimodal policy learning such as VLA fine-tuning in particular. As the whole is greater than the sum of its parts, the framework demonstrates strong performance, extensibility, and adaptability — all achieved at low computational cost and with a short development cycle of fine-tuning.

Moreover, it also suggests a potential new step change in VLA-based physical AI systems, which can be leveraged by VLMs' perceptual understanding ability to match human-level cognition, and thus to deal with complicated scenarios, where a repertoire of autonomous robot skills and human-in-the-loop mechanism can be synergistically combined for deploying such technologies in real-world industrial sectors.

\newpage
\section{Contributions \& Acknowledgements}
\label{sec:contributions}

The following contributions are categorized by technical and research inputs, and the authors' names are listed within each category.

\begin{itemize}
    \item Research Ideas and Conceptualization: Yu Cui, Zhibin Li.
    \item Teleoperation System: Yujian Zhang, Xinyu Yi.
    \item Data Collection and Curation: Yujian Zhang, Yu Cui, Yang Li, Xinyu Yi, Lina Tao.
    \item VLA Model Architecture and Innovation: Yu Cui, Lina Tao, Yujian Zhang, Yang Li, Zhibin Li.
    \item Training: Yu Cui, Yujian Zhang, Lina Tao, Yang Li.
    \item Evaluation and Analysis: Yu Cui, Yujian Zhang, Yang Li, Lina Tao. 
    \item Hardware \& Software Infrastructure: Yujian Zhang, Yu Cui, Yang Li.  
    \item Paper Writing: All authors.
    \item Research Direction and Team Lead: Zhibin Li.
\end{itemize}  

We sincerely thank Jianran Liu, Yunfei Liu, Lisuo Li for their generous and kind support on hardware setup, and Wanli Peng for his support in providing hardware accessories. We thank Heyang Xu and Ke Fan for their help on the early prototyping of the VLA systems, and Jieyi Zhang for his early development of the LSTM-based grasping policies. 

We extend our sincere gratitude to Wenjia Zhu, for encouraging the focus on high-value research. We are very thankful to Yonghui Wu for his strategic direction in shaping research priorities. Furthermore, our special acknowledgment goes to the Department Head, Hang Li, for his substantial enabling and organizational support throughout this project.

\clearpage

\beginappendix

\section{Implementation Details}
\subsection{Model Fine-Tuning}
All Vision-Language-Action (VLA) policies in this work are derived from a pre-trained base model $\pi_0$. The fine-tuning process is conducted as follows:
\begin{itemize}
\item \textbf{Parameter Update:} For both the hand-only and the arm-hand VLA policies, we perform full-parameter fine-tuning on $\pi_0$, with the exception of the visual encoder, which remains frozen.
\item \textbf{Task Instruction:} The specific language instruction provided to the model is tailored to the task:
\begin{itemize}
\item Instruction for the \textbf{hand-only VLA} policy: pick up the object on the table and place it elsewhere. 
\item Instruction for the \textbf{arm-hand VLA} policy: pick up the object on the table and place it in the box.
\end{itemize}
\end{itemize}

\subsection{Hardware Platforms and Real-World Deployment}
\label{sec:supplementary:hardware}
\begin{figure}[!h]
    \centering
    \includegraphics[width=0.9\textwidth]{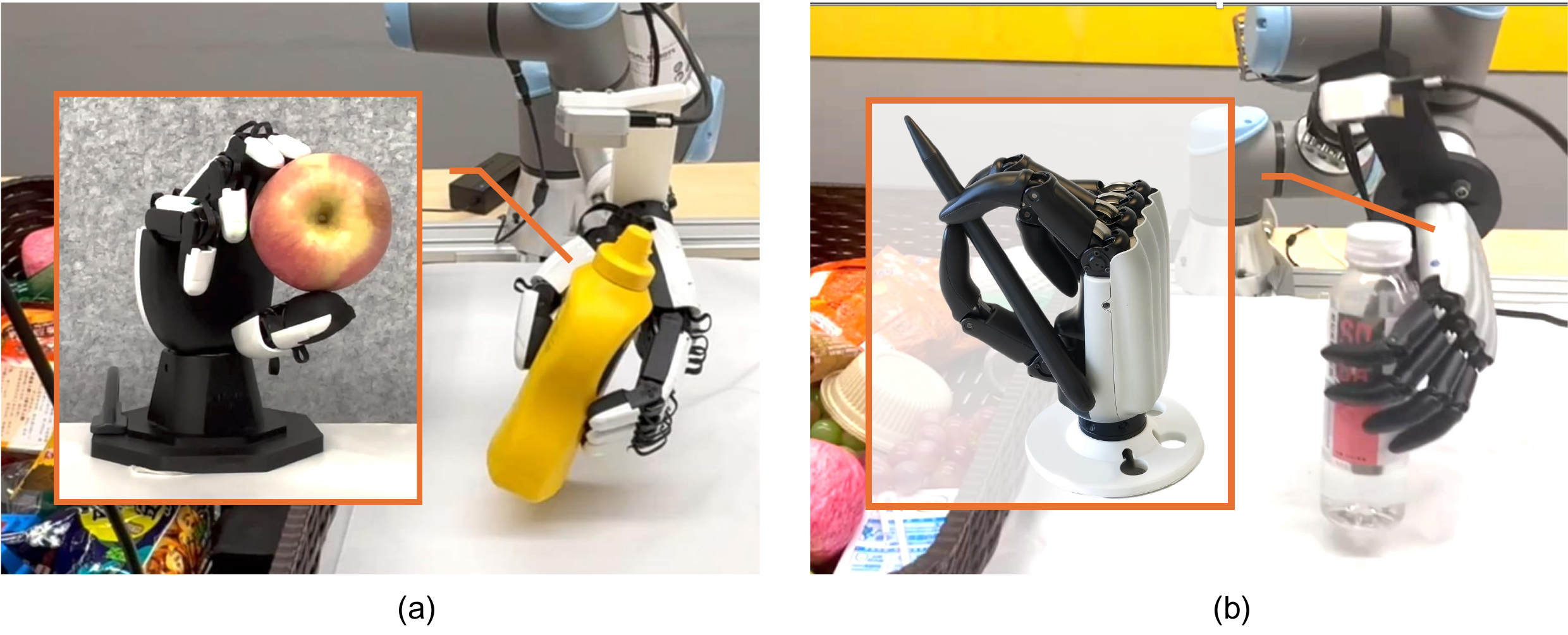}
    \vspace{-3mm}
    \caption{Cross-embodiment validation using two different robotic hands for both the Shared Autonomy framework and the training of end-to-end VLA models: (a) XHAND1 hand \cite{xhand}, and (b) RY-H2 hand \cite{ruiyan}. }
    \label{fig:hands}
    \vspace{-2mm}
\end{figure}

As the research community embraces a range of different designs of hardware, we validated the applicability of our proposed framework using two different dexterous robotic hands which are quite representative (one is fully actuated and the other is underactuated): (1) \texttt{Xhand}, a high-performance 12-DoF dexterous robotic hand with fingertip tactile sensing, as the main hardware; (2) RY-H2, a fast 11-DoF dexterous hand (6 active DoF and 5 under-actuated DoF) with joint current sensing featured by a quick open-close cycle of 0.4s. 

The \texttt{Xhand} provides a high-bandwidth motor control interface for learning-based control, joint-level state data, and high-resolution tactile feedback \cite{xhand}. Its five fingertips are equipped with \( 270^\circ \) encirclement of tactile array sensors, providing 120 channels in total for the 3-dimensional force data per fingertip, providing tactile perception of contact geometry and distributed forces. The motors offer multiple control modes, including position, force, and hybrid force-position control, running at a control frequency of 83 Hz over an EtherCAT. With a fingertip grip force of 15 \textit{N} and a maximum grip force of 80 \textit{N}, the hand can perform both precise and powerful grasps. Its design of back-drivable actuation and a lifetime of 1000000 grasp cycles, makes it suitable for the extensive trial-and-error required in data collection and testing trials of VLA policies.

The RY-H2 hand is a five-finger under-actuated dexterous hand featured by high-speed grasping \cite{ruiyan}. Totaling 11 joints, with 6 active and 5 passive degrees of freedom, it is actuated by high-power-density brushless DC motors. This design enables a rapid open-close cycle time of 0.4\textit{s} and a high maximum grip force of 140\textit{N}. With a lightweight of 0.6\textit{kg}, it is suitable for dynamic tasks requiring both speed and power, and secondary algorithmic development for industrial and research applications.

During the evaluation phase on these physical robot hardware, the same setting was configured:
\begin{itemize}
\item \textbf{Model Checkpoint:} All VLA models evaluated in the main experiments and the appendix are the checkpoints saved at 80,000 training steps.
\item \textbf{Control Frequency:} The robot arm and hand are controlled by the policy at \textbf{30 Hz} control frequency.
\end{itemize}

\subsection{Network Architecture Specifications}

The Arm-Hand Feature Enhancement module extends the base architecture with dedicated components for limb-specific feature extraction. The arm encoder $\mathcal{E}_{\text{arm}}$ and the hand encoder $\mathcal{E}_{\text{hand}}$ are both implemented as a two-layer MLP.
Each encoder takes the shared representation $z_t^{\text{share}} \in \mathbb{R}^{d_s}$ as input and produces limb-specific features of reduced dimensionality $z_t^{\text{arm}} \in \mathbb{R}^{d_s/2}$ and $z_t^{\text{hand}} \in \mathbb{R}^{d_s/2}$ through successive linear transformations separated by the Mish activation functions.

The auxiliary prediction heads $\mathcal{H}_{\text{arm}}$ and $\mathcal{H}_{\text{hand}}$ are implemented as single linear layers that map the limb-specific features to action predictions of a fixed max dimension, with selective supervision applied only to the indices corresponding to each limb's actual degrees of freedom.

\section{Additional Results}
\label{sec:app:addition}

\subsection{Effectiveness of Shared Autonomy Data Collection}
 \begin{table*}[htbp] 
  \centering
  \vskip -0.2cm 
  \small
  \setlength{\tabcolsep}{4pt} 
  \caption{Data collection efficiency and training/deploying expenditure of shared autonomy vs full teleoperation.}
  \label{tab:vla_succ}
  \begin{tabular}{c|c|c|c|c}
    \toprule
    \textbf{Methods} & Main Collection & Corrective Collection & Fine-Tuning Time & Deploying Time \\
    \midrule
    \textbf{Shared Autonomy} & 110/hour/person & 100/hour/person & 4 hours (on 4 GPUs) & 10--15 minutes (20 trials) \\ 
    \textbf{Full Teleoperation} & 90/hour/person & 80/hour/person & NA & NA \\
    \bottomrule
  \end{tabular}
  \vskip -0.2cm 
  \label{tab:shared}
\end{table*}
Our Shared Autonomy framework demonstrates a clear advantage in data collection efficiency. 
As shown in Table~\ref{tab:shared}, it allows a single operator to collect 110 trajectories per hour for the main dataset, compared to 90 with full teleoperation. This ~25\% increase in collection rate, sustained during corrective data collection (100 vs. 80 trajectories/hour), directly translates to faster policy improvement cycles and validates the framework's effectiveness. 

This high-quality data allows for efficient policy refinement: a fine-tuning run (20k steps on 4 GPUs) completes in 4 hours, and deploying the refined policy for 20 evaluation trials takes only 10–15 minutes. This end-to-end efficiency shows the feasibility of our approach for rapid policy training and iteration. For skilled operators, collecting more than 100 demos per hour per person is easily and readily achievable, and most domain-specific cases usually require around 50 demos only, which enables a development-to-deployment cycle of one day only.

\subsection{Additional Results of End-to-End Adaptive Arm-Hand Grasping}
\label{sec:grasp_more}

\begin{figure}[!h]
    \centering
    \includegraphics[width=1.0\textwidth]{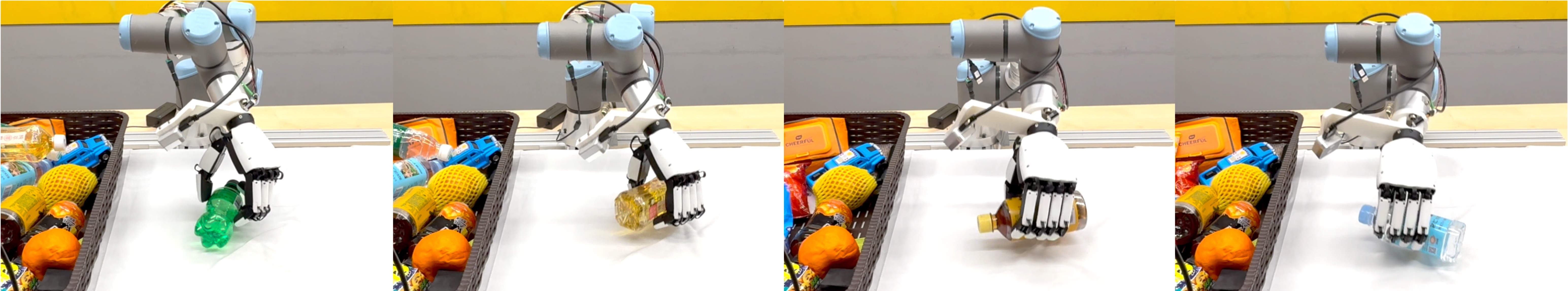}
    \caption{Grasping performance across different object orientations.}
    \label{fig:different_poses}
    \vspace{0mm}
\end{figure}

\begin{figure}[!h]
    \centering
    \includegraphics[width=1.0\textwidth]{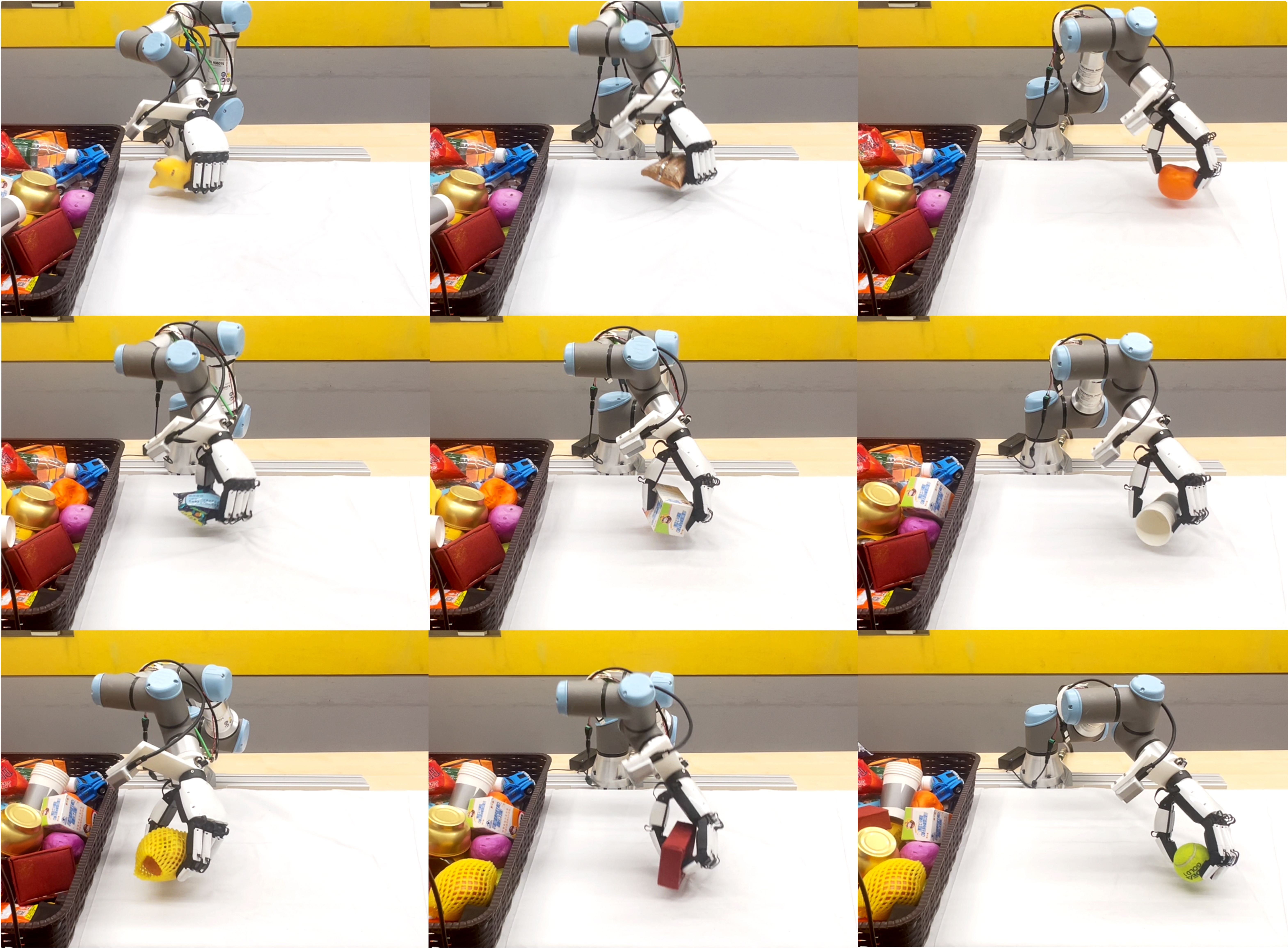}
    \caption{Grasping robustness tested at various spatial locations within the workspace.}
    \label{fig:different_locations}
    \vspace{-5mm}
\end{figure}

This section presents additional qualitative results, and Fig.~\ref{fig:different_poses} to Fig.~\ref{fig:different_locations} demonstrate the robustness and generalization capability of our end-to-end VLA policy in additional grasping scenarios.

\subsubsection{Effectiveness of Tactile Sensing in \texorpdfstring{$\pi_{\text{uni}}$}{pi(uni)}}
To evaluate the impact of tactile sensing on the full arm-hand policy, we conducted an additional experiment under the same conditions as our main \texttt{Xhand} evaluation (20×20 cm workspace, 10 objects, 10 trials each). The tactile representation and integration method followed the same approach as in the hand-only DexGrasp-VLA policy. Despite the tactile sensing provides significant benefits to the hand-only policy $\pi_{\text{hand}}$, we found that incorporating these same tactile features into the unified arm-hand policy $\pi_{\text{uni}}$ does not consistently improve performance. In other words, tactile sensing is more effective when used in combination with local visual sensing for the hand-only DexGrasp VLA policy, but the direct incorporation of tactile sensing in the unified arm-hand policy $\pi_{\text{uni}}$ does \textit{not} yield positive results, at least from this initial study. Specifically, the enhanced policy with tactile input ($\pi_{\text{uni-enhance-tac}}$) achieved a success rate of 82\%, compared to 95\% for the visual-proprioceptive only policy ($\pi_{\text{uni-enhance}}$) trained by datasets collected by shared autonomy.

This performance degradation is likely due to the different functional roles for controlling the arm and the hand respectively. To the best of our knowledge, we hypothesize that the arm primarily executes reaching motions that rely more on visual and proprioceptive feedback for spatial motions, while tactile signals are most relevant for fine-grained grasping and in-hand manipulation. Uniformly incorporating tactile input throughout the entire arm-hand trajectory may introduce irrelevant information like ``noises'' during arm movement phases, particularly from incidental environmental contacts (e.g., table collisions or unintended fingertip brushing) that occur during reaching. These transient and often misleading tactile signals appear to interfere with the policy's ability to maintain robust arm-centric coordination. 

The above results suggest that future work should explore more structured tactile integration strategies rather than uniform feature fusion throughout the entire motion. Promising directions include selective sensor gating mechanisms that activate tactile processing only during grasping phases, or attention-based architectures that learn to dynamically weight tactile input based on the current task phase. Such approaches could preserve the benefits of tactile sensing for manipulation, while avoiding the performance degradation observed during arm movements.

\subsubsection{Additional Results of Corrective Control for Refining a VLA Policy}
\label{sec:appendix_corrective}

This appendix extends the experimental validation of our corrective framework beyond the pick-and-place tasks presented in the main text. Together with the shared autonomy approach used for grasping tasks in the main experiments, the additional studies here, which employed teleoperation for long-horizon tasks and motion planning for industrial assembly, provide comprehensive evidence for the generality of our corrective mechanism across diverse tasks and data collection methodologies.

\subsection{Long-Horizon Manipulation with Robotic Gripper}
\label{sec:appendix:gripper}

\subsubsection{Long-Horizon Tasks Learned from Teleoperation}
\label{sec:long_horizon}

\begin{figure}[!h]
    \centering
    \includegraphics[width=1.0\textwidth]{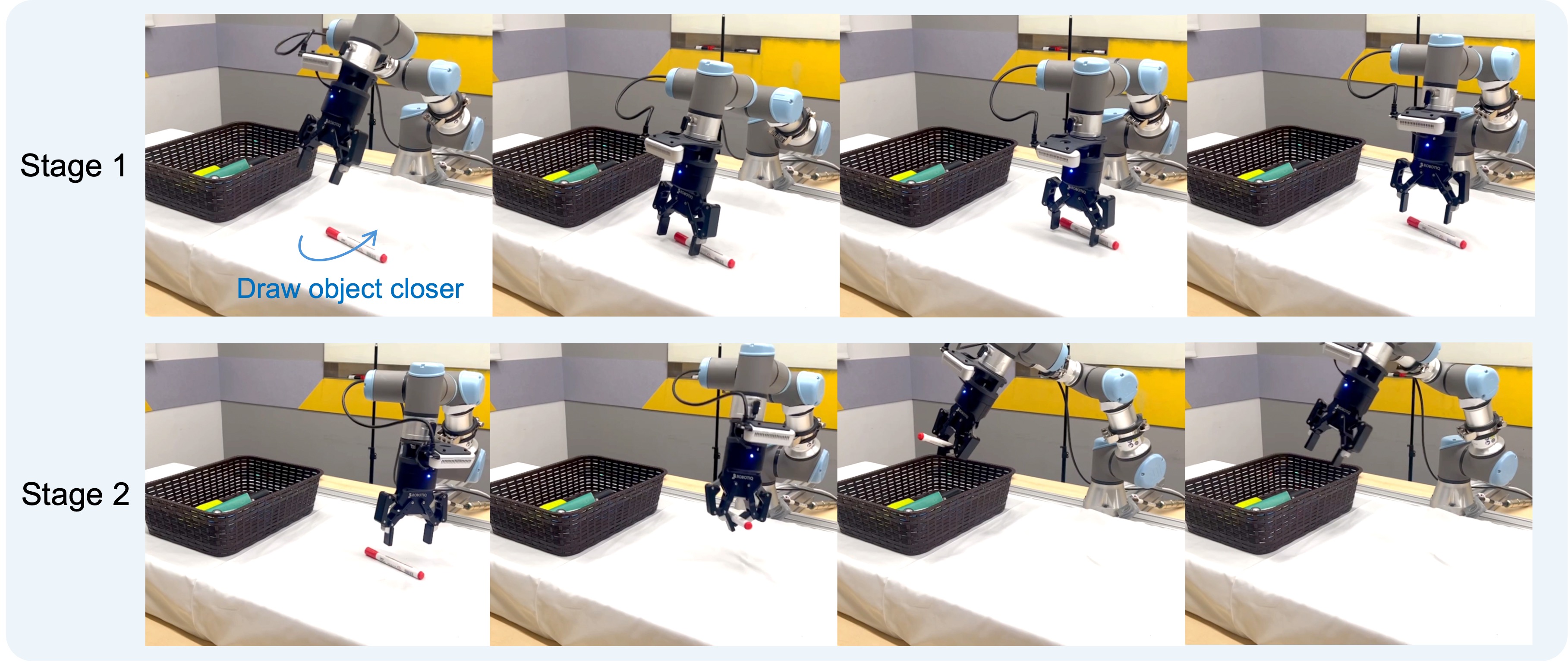}
    \caption{\normalsize \textbf{Task 1}. Sequential object manipulation via multi-stage physical interactions with a distant object. The task involves a two-stage strategy to manipulate a pen from a distant, non-graspable location (at the arm's far reach near singularity if performing the usual grasp) to a nearby location suitable for grasping. Stage 1: the robot sweeps, reorients, and draws the object closer to its base. Stage 2: the robot executes a final grasp and placement.}
    \label{fig:task1}
    \vspace{-2mm}
\end{figure}

\begin{figure}[!h]
    \centering
    \includegraphics[width=1.0\textwidth]{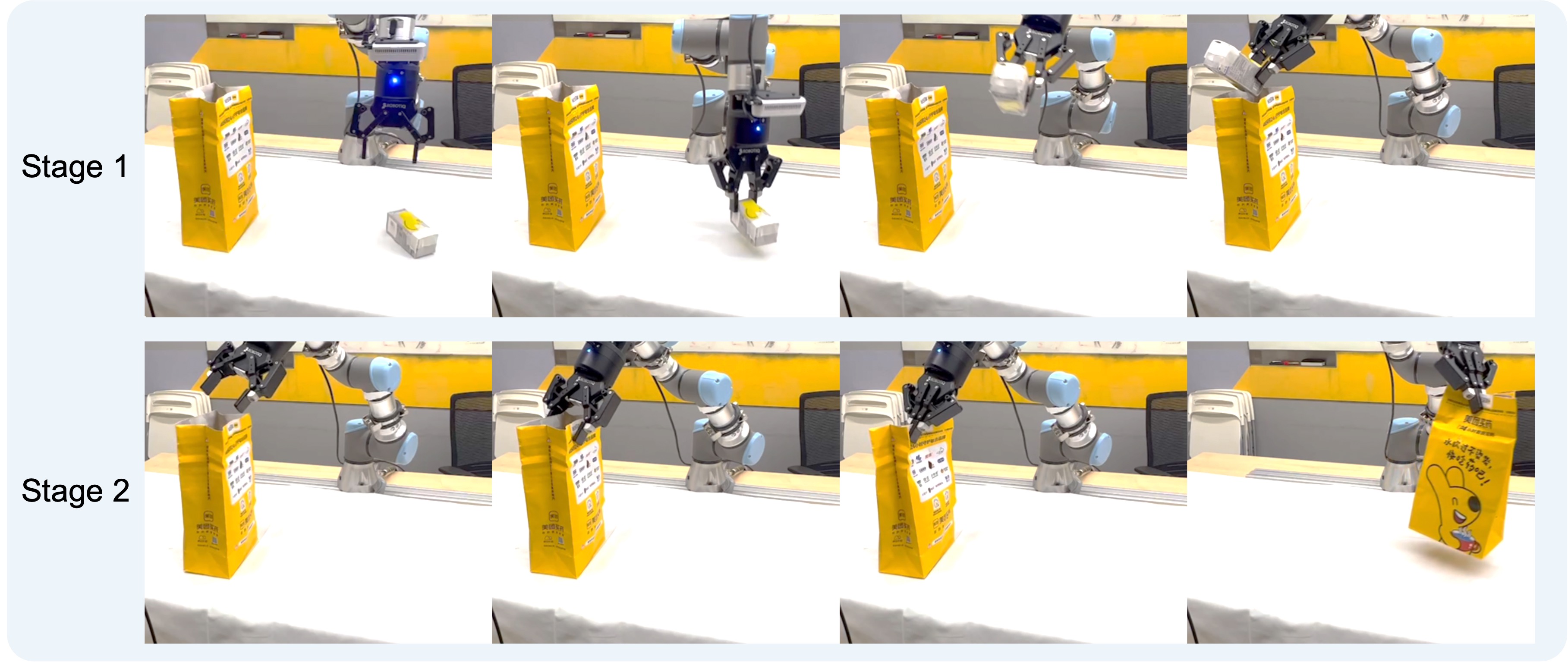}
    \caption{\normalsize \textbf{Task 2}. Sequential multi-stage packing showing key actions. Stage 1: placing the medication box into the packaging bag. Stage 2: execution of closing the bag and transporting the closed package, showing the capability of operating both rigid and deformable objects.}
    \label{fig:task2}
    \vspace{-2mm}
\end{figure}

\begin{figure}[!h]
    \centering
    \includegraphics[width=1.0\textwidth]{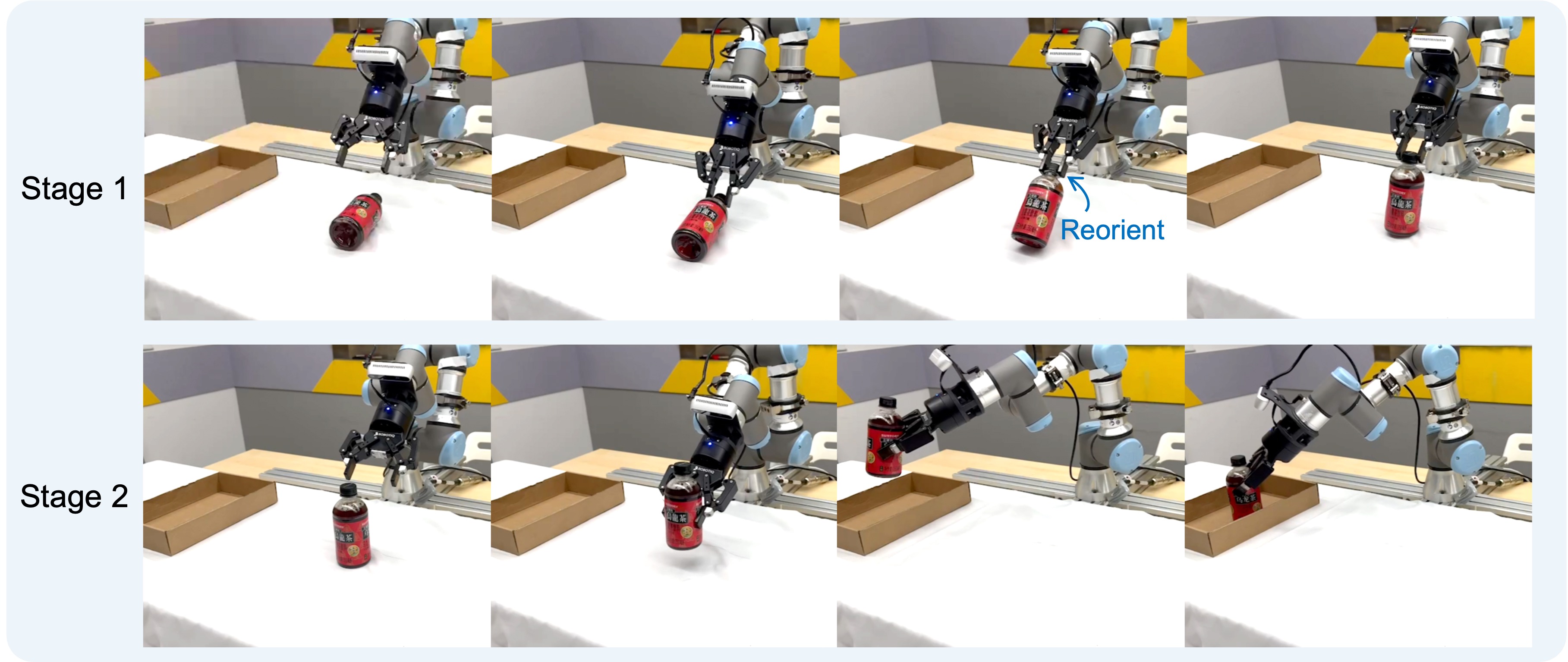}
    \caption{\normalsize \textbf{Task 3}. Underactuated bottle uprighting by exploiting low torsional friction. Stage 1: grasping the cap and lifting the bottle to induce passive uprighting via low torsional friction around the cap. Stage 2: subsequent pick-and-place from a re-configured graspable pose to the target box.}
    \label{fig:task3}
    \vspace{-5mm}
\end{figure}

Building upon the corrective framework validated in the main text's using dexterous hands, we further designed three sequential long-horizon tasks using parallel jaw grippers to evaluate the framework's efficacy under different hardware and task settings. These three tasks include:

\textbf{Task 1. Pen Relocation and Placement} (Fig.~\ref{fig:task1}): This task requires the robot to first sweep a distant slender pen and bring it to a closer range, then to reorient the gripper to an appropriate angle, grasp and place the pen. Common failure modes include inadequate sweeping force and incorrect gripper orientation during the pre-grasp phase. In this scenario, the robot needs to first re-configure the object and change it from a non-graspable state into a graspable pose, which requires the design of RL policies previously with handcrafted reward design~\cite{sun2020learning} and now can be learned through a VLA-based imitation by providing demonstrations for learning such behaviors. 

\textbf{Task 2. Pill Box Packing} (Fig.~\ref{fig:task2}): In this multi-stage task, the robot must sequentially grasp a small medication box, put it into a packaging bag, securely close the bag, and finally transport the entire package to a designated location. Failures typically occur during the delicate bag-closing phase and when handling the combined object during transport.

\textbf{Task 3. Bottle Uprighting and Placement} (Fig.~\ref{fig:task3}): This challenging task begins with re-configuring the state of the object -- grasping the cap of a horizontally positioned water bottle, followed by carefully lifting and uprighting the bottle into a vertical orientation -- and then completes with a standard pick-and-place operation. The uprighting process presents particular difficulties in maintaining a stable grip, taking advantage of the low torsional friction for passive rotation of the bottle during the execution of the lifting trajectory.

Following our established framework, we first trained a base policy $\pi_{\text{base}}$ on initial demonstrations. Then, we collected corrective trajectories via human teleoperation for those occurred failures across these diverse task scenarios. After fine-tuning, the resulting policy $\pi_{\text{corr}}$ achieved significantly higher success rates (see Table~\ref{tab:long-horizon}), demonstrating that the corrective mechanism (as previously shown effective with shared autonomy) also works well for the hardware system using teleoperation in complex multi-step tasks with parallel jaw grippers.

 \begin{table*}[htbp]
  \centering
  \small
  \setlength{\tabcolsep}{8pt}
  \caption{Success rates of long-horizon tasks via corrective teleoperation.}
  \label{tab:vla_succ}
  \begin{tabular}{c|c|c|c}
    \toprule
    \textbf{Tasks} & \textbf{Task 1} (Fig.~\ref{fig:task1}) & \textbf{Task 2} (Fig.~\ref{fig:task2}) & \textbf{Task 3} (Fig.~\ref{fig:task3}) \\
    \midrule 
    Success Rates & 65\% & 90\% & 70\% \\
    \bottomrule
  \end{tabular}
  \label{tab:long-horizon}
  \vskip -0.2cm 
\end{table*}


\subsubsection{Learning Industrial Assembly Task - Data Collection through Motion Planning}
\label{sec:assembly}

\begin{figure}[!h]
\centering
\includegraphics[width=1.0\textwidth]{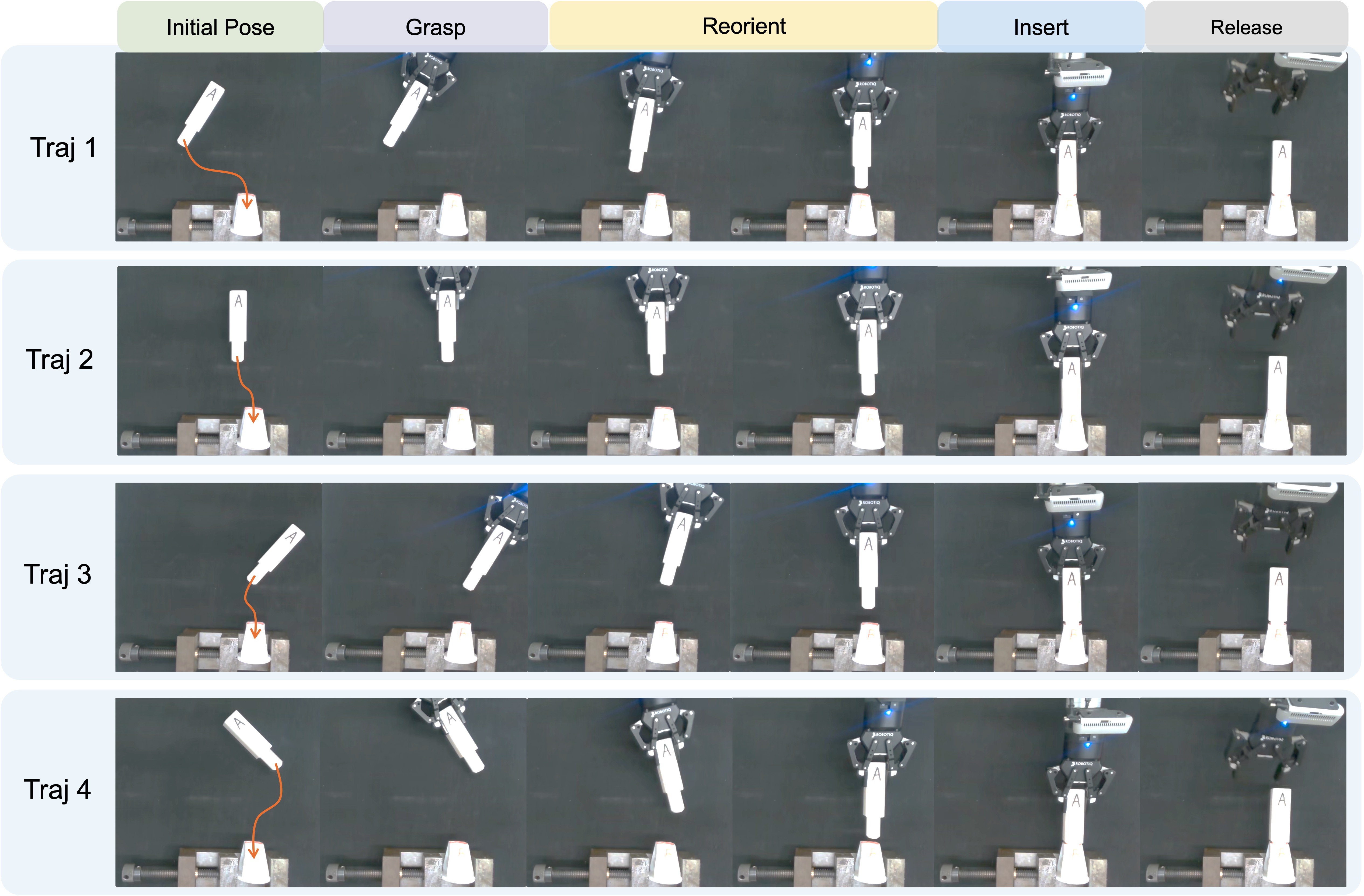}
\caption{Peg-in-hole assembly task showing multiple stages (grasp, reorient, insert, release) under four different initial configurations.}
\label{fig:peg_assembly}
\end{figure}

\begin{figure}[!h]
\centering
\includegraphics[width=1.0\textwidth]{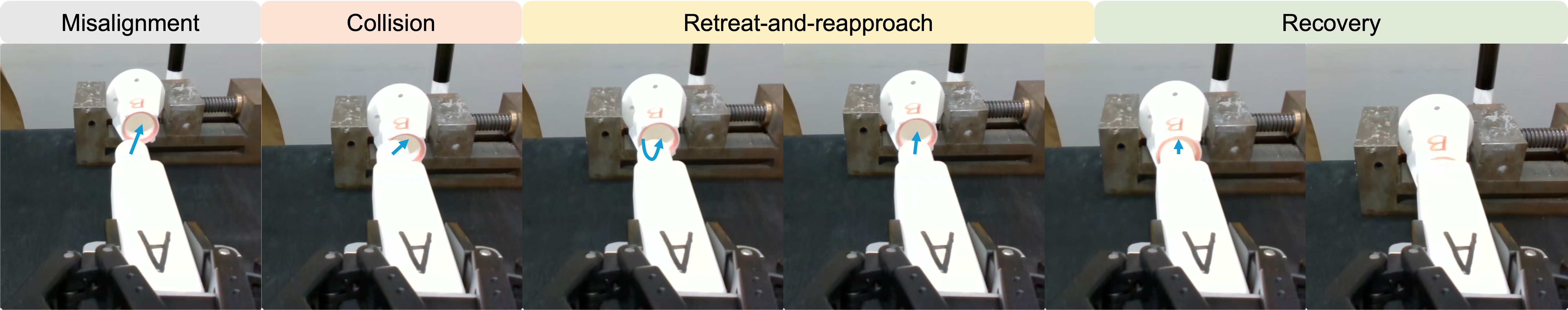}
\caption{Time-elapsed frames of error-recovery during peg-in-hole assembly, showing stages of misalignment, collision, retreat-and-reapproach, and failure recovery with successful insertion.}
\label{fig:recovery_stages}
\vspace{-3mm}
\end{figure}

Complementing the human-in-the-loop approaches (such as the shared autonomy in the main text and the teleoperation in long-horizon tasks in the Appendix), we investigated how \textit{automated motion planning} could generate effective failure-correction and serve as a useful source of data. We applied our framework to a peg-in-hole assembly task -- a canonical industrial task that requires spatial accuracy beyond typical pick-and-place. Following the same experimental protocol, we began with 100 initial demonstrations to train $\pi_{\text{base}}$ using motion planning, where the representative cases from data collection are shown in Fig.~\ref{fig:peg_assembly}. 

\begin{wraptable}{r}{0.3\linewidth} 
  \centering
  \vspace{0mm} 
  \small
  \setlength{\tabcolsep}{6pt} 
  \caption{Success rates of the task of peg-in-hole. }
  \label{tab:vla_succ}
  \begin{tabular}{c|c|c}
    \toprule
    \textbf{Methods} & \textbf{$\pi_{\text{base}}$} & \textbf{$\pi_{\text{corr}}$} \\
    \midrule
    Peg-in-Hole & 70\% & 90\%  \\
    \bottomrule
  \end{tabular}
  \label{tab:peg-in-hole}
  \vspace{-5mm} 
\end{wraptable}

When the policy failed on precision-critical cases, we employed \textbf{automated motion planning} to generate corrective trajectories (Fig.~\ref{fig:recovery_stages}), instead of human intervention. The motion planner produced recovery behaviors, including fine adjustments and retreat-and-reapproach motions. Reported as in Table~\ref{tab:peg-in-hole}, after incorporating 20 additional recovery trajectories and fine-tuning, $\pi_{\text{corr}}$ achieved a 90\% success rate: \textbf{20\% absolute improvement} over $\pi_{\text{base}}$. This result confirms that the corrective mechanism works effectively even with fully automated correction and its generated data sources, extending its applicability beyond human-guided interventions.

\subsection{Discussion and Summary}
The collective evidence from all experiments -- spanning pick-and-place tasks with shared autonomy (main text), long-horizon tasks with teleoperation, and industrial assembly with motion planning -- consistently demonstrates the effectiveness of our corrective framework across task types and data collection approaches. Each study followed the same fundamental principle: initial policy training, failure identification, corrective data collection, and policy re-training/refinement, while employing different failure-correction approaches and data sources tailored to the task-specific requirements. This iterative approach, despite relying on manual intervention and human involvement, represents a preliminary implementation of a closed-loop data flywheel that continuously retrains our VLA models with real-world interactions and the correspondingly generated real-robot data.

This progression from human-guided corrections (e.g., shared autonomy, teleoperation) to fully automated solutions (e.g., motion planning) highlights the framework's core versatility, which can be tailored to different functional components for a wide range of tasks. The shared autonomy approach balances human expertise with automated assistance for efficient grasping corrections; teleoperation provides full human control for complex multi-step tasks; while motion planning offers a fully automated solution for structured industrial environments.

Most importantly, all three approaches yield significant performance improvements, confirming that the corrective mechanism itself is the key driver of policy refinement and enhancement, while any specific data collection method merely serves as a means to this end. This inherent versatility makes our framework applicable across a broad spectrum of robotic learning scenarios, from human-centric environments to structured industrial settings. By providing an effective means for continuous improvement, our work paves the way for more capable and robust general-purpose VLA policies, ultimately expanding the reach of advanced autonomous systems in the real world.

\clearpage
\bibliographystyle{plainnat}
\bibliography{main}

\begin{thebibliography}{57}
\providecommand{\natexlab}[1]{#1}
\providecommand{\url}[1]{\texttt{#1}}
\expandafter\ifx\csname urlstyle\endcsname\relax
  \providecommand{\doi}[1]{doi: #1}\else
  \providecommand{\doi}{doi: \begingroup \urlstyle{rm}\Url}\fi

\bibitem[rui()]{ruiyan}
{RY-H2}.
\newblock URL \url{http://www.ruiyanrobot.com/product/hand/35}.

\bibitem[xha()]{xhand}
{XHAND1}.
\newblock URL \url{https://www.robotera.com/en/goods1/4.html}.

\bibitem[Bi et~al.(2025)Bi, Ma, Hao, Shou, and Soh]{bi2025vla-touch}
Jianxin Bi, Kevin~Yuchen Ma, Ce~Hao, Mike~Zheng Shou, and Harold Soh.
\newblock Vla-touch: Enhancing vision-language-action models with dual-level tactile feedback.
\newblock \emph{arXiv:2507.17294}, 2025.

\bibitem[Bjorck et~al.(2025)Bjorck, Casta{\~n}eda, Cherniadev, Da, Ding, Fan, Fang, Fox, Hu, Huang, et~al.]{bjorck2025gr00t}
Johan Bjorck, Fernando Casta{\~n}eda, Nikita Cherniadev, Xingye Da, Runyu Ding, Linxi Fan, Yu~Fang, Dieter Fox, Fengyuan Hu, Spencer Huang, et~al.
\newblock Gr00t n1: An open foundation model for generalist humanoid robots.
\newblock \emph{arXiv:2503.14734}, 2025.

\bibitem[Black et~al.()Black, Brown, Driess, Esmail, Equi, Finn, Fusai, Groom, Hausman, Ichter, et~al.]{black2410pi0}
Kevin Black, Noah Brown, Danny Driess, Adnan Esmail, Michael Equi, Chelsea Finn, Niccolo Fusai, Lachy Groom, Karol Hausman, Brian Ichter, et~al.
\newblock $\pi$0: A vision-language-action flow model for general robot control. corr, abs/2410.24164, 2024. doi: 10.48550.
\newblock \emph{arXiv.2410.24164}.

\bibitem[Bu et~al.(2025)Bu, Cai, Chen, Cui, Ding, Feng, Gao, He, Hu, Huang, et~al.]{bu2025agibot}
Qingwen Bu, Jisong Cai, Li~Chen, Xiuqi Cui, Yan Ding, Siyuan Feng, Shenyuan Gao, Xindong He, Xuan Hu, Xu~Huang, et~al.
\newblock Agibot world colosseo: A large-scale manipulation platform for scalable and intelligent embodied systems.
\newblock \emph{arXiv:2503.06669}, 2025.

\bibitem[Cadene et~al.(2024)Cadene, Alibert, Soare, Gallouedec, Zouitine, Palma, Kooijmans, Aractingi, Shukor, Aubakirova, Russi, Capuano, Pascal, Choghari, Moss, and Wolf]{cadene2024lerobot}
Remi Cadene, Simon Alibert, Alexander Soare, Quentin Gallouedec, Adil Zouitine, Steven Palma, Pepijn Kooijmans, Michel Aractingi, Mustafa Shukor, Dana Aubakirova, Martino Russi, Francesco Capuano, Caroline Pascal, Jade Choghari, Jess Moss, and Thomas Wolf.
\newblock Lerobot: State-of-the-art machine learning for real-world robotics in pytorch.
\newblock \url{https://github.com/huggingface/lerobot}, 2024.

\bibitem[Carpentier et~al.(2019)Carpentier, Saurel, Buondonno, Mirabel, Lamiraux, Stasse, and Mansard]{carpentier2019pinocchio}
Justin Carpentier, Guilhem Saurel, Gabriele Buondonno, Joseph Mirabel, Florent Lamiraux, Olivier Stasse, and Nicolas Mansard.
\newblock The pinocchio c++ library -- a fast and flexible implementation of rigid body dynamics algorithms and their analytical derivatives.
\newblock In \emph{IEEE International Symposium on System Integrations (SII)}, 2019.

\bibitem[Chen et~al.(2025)Chen, Chen, Chen, Cai, Liu, Liang, Li, Lin, Ge, Gu, et~al.]{chen2025robotwin2}
Tianxing Chen, Zanxin Chen, Baijun Chen, Zijian Cai, Yibin Liu, Qiwei Liang, Zixuan Li, Xianliang Lin, Yiheng Ge, Zhenyu Gu, et~al.
\newblock Robotwin 2.0: A scalable data generator and benchmark with strong domain randomization for robust bimanual robotic manipulation.
\newblock \emph{arXiv preprint arXiv:2506.18088}, 2025.

\bibitem[Chen et~al.(2023)Chen, Geng, Zhong, Ji, Jiang, Lu, Dong, and Yang]{chen2023bi}
Yuanpei Chen, Yiran Geng, Fangwei Zhong, Jiaming Ji, Jiechuang Jiang, Zongqing Lu, Hao Dong, and Yaodong Yang.
\newblock Bi-dexhands: Towards human-level bimanual dexterous manipulation.
\newblock \emph{IEEE Transactions on Pattern Analysis and Machine Intelligence}, 46\penalty0 (5):\penalty0 2804--2818, 2023.

\bibitem[Cheng et~al.(2024)Cheng, Li, Yang, Yang, and Wang]{cheng2024open}
Xuxin Cheng, Jialong Li, Shiqi Yang, Ge~Yang, and Xiaolong Wang.
\newblock Open-television: Teleoperation with immersive active visual feedback.
\newblock \emph{arXiv:2407.01512}, 2024.

\bibitem[Cheng et~al.(2025)Cheng, Zhang, Zhang, Li, Wang, Song, and Zhang]{cheng2025omnivtla}
Zhengxue Cheng, Yiqian Zhang, Wenkang Zhang, Haoyu Li, Keyu Wang, Li~Song, and Hengdi Zhang.
\newblock Omnivtla: Vision-tactile-language-action model with semantic-aligned tactile sensing.
\newblock \emph{arXiv:2508.08706}, 2025.

\bibitem[Curtis et~al.(2025)Curtis, Kumar, Cao, Lozano-P{\'e}rez, and Kaelbling]{curtis2025trust}
Aidan Curtis, Nishanth Kumar, Jing Cao, Tom{\'a}s Lozano-P{\'e}rez, and Leslie~Pack Kaelbling.
\newblock {Trust the PRoC3S: Solving Long-Horizon Robotics Problems with LLMs and Constraint Satisfaction}.
\newblock In \emph{Conference on Robot Learning}, pages 1362--1383. PMLR, 2025.

\bibitem[Deng et~al.(2025)Deng, Yan, Wei, Ma, Yang, Chen, Zhang, Yang, Zhang, Cui, et~al.]{deng2025graspvla}
Shengliang Deng, Mi~Yan, Songlin Wei, Haixin Ma, Yuxin Yang, Jiayi Chen, Zhiqi Zhang, Taoyu Yang, Xuheng Zhang, Heming Cui, et~al.
\newblock Graspvla: a grasping foundation model pre-trained on billion-scale synthetic action data.
\newblock \emph{arXiv:2505.03233}, 2025.

\bibitem[Ding et~al.(2024)Ding, Qin, Zhu, Jia, Yang, Yang, Qi, and Wang]{ding2024bunny}
Runyu Ding, Yuzhe Qin, Jiyue Zhu, Chengzhe Jia, Shiqi Yang, Ruihan Yang, Xiaojuan Qi, and Xiaolong Wang.
\newblock Bunny-visionpro: Real-time bimanual dexterous teleoperation for imitation learning.
\newblock \emph{arXiv:2407.03162}, 2024.

\bibitem[Duan et~al.(2024)Duan, Yuan, Pumacay, Wang, Ehsani, Fox, and Krishna]{duan2024manipulate-any}
Jiafei Duan, Wentao Yuan, Wilbert Pumacay, Yi~Ru Wang, Kiana Ehsani, Dieter Fox, and Ranjay Krishna.
\newblock Manipulate-anything: Automating real-world robots using vision-language models.
\newblock \emph{arXiv:2406.18915}, 2024.

\bibitem[Geng et~al.(2025)Geng, Wang, Wei, Li, Wang, An, Cheng, Lou, Li, Wang, et~al.]{geng2025roboverse}
Haoran Geng, Feishi Wang, Songlin Wei, Yuyang Li, Bangjun Wang, Boshi An, Charlie~Tianyue Cheng, Haozhe Lou, Peihao Li, Yen-Jen Wang, et~al.
\newblock Roboverse: Towards a unified platform, dataset and benchmark for scalable and generalizable robot learning.
\newblock \emph{arXiv preprint arXiv:2504.18904}, 2025.

\bibitem[Hansen et~al.(2022)Hansen, Hogan, Rivkin, Meger, Jenkin, and Dudek]{hansen2022visuotactile}
Johanna Hansen, Francois Hogan, Dmitriy Rivkin, David Meger, Michael Jenkin, and Gregory Dudek.
\newblock Visuotactile-rl: Learning multimodal manipulation policies with deep reinforcement learning.
\newblock In \emph{IEEE International Conference on Robotics and Automation}, pages 8298--8304, 2022.

\bibitem[Hu et~al.(2025)Hu, Huang, Lee, Yang, Zheng, and Li]{hu2025vt-dexterous}
Wenbin Hu, Bidan Huang, Wang~Wei Lee, Sicheng Yang, Yu~Zheng, and Zhibin Li.
\newblock Dexterous in-hand manipulation of slender cylindrical objects through deep reinforcement learning with tactile sensing.
\newblock \emph{Robotics and Autonomous Systems}, 186:\penalty0 104904, 2025.

\bibitem[Huang et~al.(2024{\natexlab{a}})Huang, Wang, Yang, Luo, and Li]{huang20243dvitac}
Binghao Huang, Yixuan Wang, Xinyi Yang, Yiyue Luo, and Yunzhu Li.
\newblock {3D-ViTac: Learning fine-grained manipulation with visuo-tactile sensing}.
\newblock \emph{arXiv:2410.24091}, 2024{\natexlab{a}}.

\bibitem[Huang et~al.(2025)Huang, Wang, Lin, Hu, Wen, and Gao]{huang2025tactile-vla}
Jialei Huang, Shuo Wang, Fanqi Lin, Yihang Hu, Chuan Wen, and Yang Gao.
\newblock Tactile-vla: Unlocking vision-language-action model's physical knowledge for tactile generalization.
\newblock \emph{arXiv:2507.09160}, 2025.

\bibitem[Huang et~al.(2024{\natexlab{b}})Huang, Wang, Li, Zhang, and Fei-Fei]{huang2024rekep}
Wenlong Huang, Chen Wang, Yunzhu Li, Ruohan Zhang, and Li~Fei-Fei.
\newblock Rekep: Spatio-temporal reasoning of relational keypoint constraints for robotic manipulation.
\newblock \emph{arXiv:2409.01652}, 2024{\natexlab{b}}.

\bibitem[Intelligence et~al.()Intelligence, Black, Brown, Darpinian, Dhabalia, Driess, Esmail, Equi, Finn, Fusai, et~al.]{intelligence2504pi0}
Physical Intelligence, Kevin Black, Noah Brown, James Darpinian, Karan Dhabalia, Danny Driess, Adnan Esmail, Michael Equi, Chelsea Finn, Niccolo Fusai, et~al.
\newblock $\pi$0. 5: a vision-language-action model with open-world generalization, 2025.
\newblock \emph{URL https://arxiv. org/abs/2504.16054}, 1\penalty0 (2):\penalty0 3.

\bibitem[Khazatsky et~al.(2024)Khazatsky, Pertsch, Nair, Balakrishna, Dasari, Karamcheti, Nasiriany, Srirama, Chen, Ellis, et~al.]{khazatsky2024droid}
Alexander Khazatsky, Karl Pertsch, Suraj Nair, Ashwin Balakrishna, Sudeep Dasari, Siddharth Karamcheti, Soroush Nasiriany, Mohan~Kumar Srirama, Lawrence~Yunliang Chen, Kirsty Ellis, et~al.
\newblock Droid: A large-scale in-the-wild robot manipulation dataset.
\newblock \emph{arXiv:2403.12945}, 2024.

\bibitem[Kim et~al.(2024)Kim, Pertsch, Karamcheti, Xiao, Balakrishna, Nair, Rafailov, Foster, Lam, Sanketi, et~al.]{kim2024openvla}
Moo~Jin Kim, Karl Pertsch, Siddharth Karamcheti, Ted Xiao, Ashwin Balakrishna, Suraj Nair, Rafael Rafailov, Ethan Foster, Grace Lam, Pannag Sanketi, et~al.
\newblock Openvla: An open-source vision-language-action model.
\newblock \emph{arXiv:2406.09246}, 2024.

\bibitem[Kumar et~al.(2024)Kumar, Shen, Ramos, Fox, Lozano-P{\'e}rez, Kaelbling, and Garrett]{kumar2024open}
Nishanth Kumar, William Shen, Fabio Ramos, Dieter Fox, Tom{\'a}s Lozano-P{\'e}rez, Leslie~Pack Kaelbling, and Caelan~Reed Garrett.
\newblock Open-world task and motion planning via vision-language model inferred constraints.
\newblock \emph{arXiv:2411.08253}, 2024.

\bibitem[Lee et~al.(2020)Lee, Zhu, Zachares, Tan, Srinivasan, Savarese, Fei-Fei, Garg, and Bohg]{lee2020making}
Michelle~A Lee, Yuke Zhu, Peter Zachares, Matthew Tan, Krishnan Srinivasan, Silvio Savarese, Li~Fei-Fei, Animesh Garg, and Jeannette Bohg.
\newblock Making sense of vision and touch: Learning multimodal representations for contact-rich tasks.
\newblock \emph{IEEE Transactions on Robotics}, 36\penalty0 (3):\penalty0 582--596, 2020.

\bibitem[Lin et~al.(2024)Lin, Zhang, Li, Qi, Yi, Levine, and Malik]{lin2024visuotactile}
Toru Lin, Yu~Zhang, Qiyang Li, Haozhi Qi, Brent Yi, Sergey Levine, and Jitendra Malik.
\newblock Learning visuotactile skills with two multifingered hands.
\newblock \emph{arXiv:2404.16823}, 2024.

\bibitem[Lin et~al.(2025)Lin, Sachdev, Fan, Malik, and Zhu]{lin2025rlsim2real}
Toru Lin, Kartik Sachdev, Linxi Fan, Jitendra Malik, and Yuke Zhu.
\newblock Sim-to-real reinforcement learning for vision-based dexterous manipulation on humanoids.
\newblock \emph{arXiv:2502.20396}, 2025.

\bibitem[Liu et~al.(2025{\natexlab{a}})Liu, Li, Qin, Shaw, Xu, Abbeel, and Chen]{liu2025vitamin}
Fangchen Liu, Chuanyu Li, Yihua Qin, Ankit Shaw, Jing Xu, Pieter Abbeel, and Rui Chen.
\newblock Vitamin: Learning contact-rich tasks through robot-free visuo-tactile manipulation interface.
\newblock \emph{arXiv:2504.06156}, 2025{\natexlab{a}}.

\bibitem[Liu et~al.(2025{\natexlab{b}})Liu, Chen, An, Liu, Zhang, Gu, Li, Guo, Chen, Liu, et~al.]{liu2025hybridvla}
Jiaming Liu, Hao Chen, Pengju An, Zhuoyang Liu, Renrui Zhang, Chenyang Gu, Xiaoqi Li, Ziyu Guo, Sixiang Chen, Mengzhen Liu, et~al.
\newblock Hybridvla: Collaborative diffusion and autoregression in a unified vision-language-action model.
\newblock \emph{arXiv:2503.10631}, 2025{\natexlab{b}}.

\bibitem[Liu et~al.(2025{\natexlab{c}})Liu, Zhang, Wang, Yu, Cui, Li, Hu, Xiong, Lu, and Wang]{aisearch2025}
Lilu Liu, Jingyu Zhang, Fei Wang, Jiyu Yu, Yuxiang Cui, Zhibin Li, Jian Hu, Rong Xiong, Haojian Lu, and Yue Wang.
\newblock {AI search, physician removal: Bronchoscopy robot bridges collaboration in foreign body aspiration}.
\newblock \emph{Science Robotics}, 10\penalty0 (104):\penalty0 eadt5338, 2025{\natexlab{c}}.

\bibitem[Liu et~al.()Liu, Cui, Sun, Li, Chen, and Ye]{liuvtdexmanip}
Qingtao Liu, Yu~Cui, Zhengnan Sun, Gaofeng Li, Jiming Chen, and Qi~Ye.
\newblock Vtdexmanip: A dataset and benchmark for visual-tactile pretraining and dexterous manipulation with reinforcement learning.
\newblock In \emph{The Thirteenth International Conference on Learning Representations}.

\bibitem[Luo et~al.(2025)Luo, Feng, Zhang, Zheng, Wang, Yuan, Liu, Xu, Jin, and Lu]{luo2025being}
Hao Luo, Yicheng Feng, Wanpeng Zhang, Sipeng Zheng, Ye~Wang, Haoqi Yuan, Jiazheng Liu, Chaoyi Xu, Qin Jin, and Zongqing Lu.
\newblock Being-h0: vision-language-action pretraining from large-scale human videos.
\newblock \emph{arXiv preprint arXiv:2507.15597}, 2025.

\bibitem[Maddukuri et~al.(2025)Maddukuri, Jiang, Chen, Nasiriany, Xie, Fang, Huang, Wang, Xu, Chernyadev, et~al.]{maddukuri2025simreal}
Abhiram Maddukuri, Zhenyu Jiang, Lawrence~Yunliang Chen, Soroush Nasiriany, Yuqi Xie, Yu~Fang, Wenqi Huang, Zu~Wang, Zhenjia Xu, Nikita Chernyadev, et~al.
\newblock Sim-and-real co-training: A simple recipe for vision-based robotic manipulation.
\newblock \emph{arXiv preprint arXiv:2503.24361}, 2025.

\bibitem[Mao et~al.(2024)Mao, Giudici, Coppola, Althoefer, Farkhatdinov, Li, and Jamone]{mao2024dexskills}
Xiaofeng Mao, Gabriele Giudici, Claudio Coppola, Kaspar Althoefer, Ildar Farkhatdinov, Zhibin Li, and Lorenzo Jamone.
\newblock {DexSkills: Skill segmentation using haptic data for learning autonomous long-horizon robotic manipulation tasks}.
\newblock In \emph{2024 IEEE/RSJ International Conference on Intelligent Robots and Systems (IROS)}, pages 5104--5111. IEEE, 2024.

\bibitem[Mu et~al.(2025)Mu, Chen, Chen, Peng, Lan, Gao, Liang, Yu, Zou, Xu, et~al.]{mu2025robotwin}
Yao Mu, Tianxing Chen, Zanxin Chen, Shijia Peng, Zhiqian Lan, Zeyu Gao, Zhixuan Liang, Qiaojun Yu, Yude Zou, Mingkun Xu, et~al.
\newblock Robotwin: Dual-arm robot benchmark with generative digital twins.
\newblock In \emph{Proceedings of the Computer Vision and Pattern Recognition Conference}, pages 27649--27660, 2025.

\bibitem[Nasiriany et~al.(2024)Nasiriany, Maddukuri, Zhang, Parikh, Lo, Joshi, Mandlekar, and Zhu]{nasiriany2024robocasa}
Soroush Nasiriany, Abhiram Maddukuri, Lance Zhang, Adeet Parikh, Aaron Lo, Abhishek Joshi, Ajay Mandlekar, and Yuke Zhu.
\newblock Robocasa: Large-scale simulation of everyday tasks for generalist robots.
\newblock \emph{arXiv preprint arXiv:2406.02523}, 2024.

\bibitem[O’Neill et~al.(2024)O’Neill, Rehman, Maddukuri, Gupta, Padalkar, Lee, Pooley, Gupta, Mandlekar, Jain, et~al.]{o2024open}
Abby O’Neill, Abdul Rehman, Abhiram Maddukuri, Abhishek Gupta, Abhishek Padalkar, Abraham Lee, Acorn Pooley, Agrim Gupta, Ajay Mandlekar, Ajinkya Jain, et~al.
\newblock Open x-embodiment: Robotic learning datasets and rt-x models: Open x-embodiment collaboration 0.
\newblock In \emph{2024 IEEE International Conference on Robotics and Automation}, pages 6892--6903. IEEE, 2024.

\bibitem[Pan et~al.(2025)Pan, Zhang, Wu, Zhao, Gao, and Dong]{pan2025omnimanip}
Mingjie Pan, Jiyao Zhang, Tianshu Wu, Yinghao Zhao, Wenlong Gao, and Hao Dong.
\newblock Omnimanip: Towards general robotic manipulation via object-centric interaction primitives as spatial constraints.
\newblock In \emph{Proceedings of the Computer Vision and Pattern Recognition Conference}, pages 17359--17369, 2025.

\bibitem[Patel et~al.(2025)Patel, Yin, Huang, Garg, Nayyeri, Fei-Fei, Lazebnik, and Li]{patel2025real}
Shivansh Patel, Xinchen Yin, Wenlong Huang, Shubham Garg, Hooshang Nayyeri, Li~Fei-Fei, Svetlana Lazebnik, and Yunzhu Li.
\newblock A real-to-sim-to-real approach to robotic manipulation with vlm-generated iterative keypoint rewards.
\newblock \emph{arXiv:2502.08643}, 2025.

\bibitem[Qin et~al.(2023)Qin, Yang, Huang, Van~Wyk, Su, Wang, Chao, and Fox]{qin2023anyteleop}
Yuzhe Qin, Wei Yang, Binghao Huang, Karl Van~Wyk, Hao Su, Xiaolong Wang, Yu-Wei Chao, and Dieter Fox.
\newblock Anyteleop: A general vision-based dexterous robot arm-hand teleoperation system.
\newblock \emph{arXiv:2307.04577}, 2023.

\bibitem[Rajeswaran et~al.(2017)Rajeswaran, Kumar, Gupta, Vezzani, Schulman, Todorov, and Levine]{rajeswaran2017learning}
Aravind Rajeswaran, Vikash Kumar, Abhishek Gupta, Giulia Vezzani, John Schulman, Emanuel Todorov, and Sergey Levine.
\newblock Learning complex dexterous manipulation with deep reinforcement learning and demonstrations.
\newblock \emph{arXiv:1709.10087}, 2017.

\bibitem[Rhoban(2025)]{placoweb}
Rhoban.
\newblock Placo: Rhoban planning and control.
\newblock https://github.com/Rhoban/placo, 2025.

\bibitem[Sun et~al.(2020)Sun, Yuan, Hu, Yang, and Li]{sun2020learning}
Zhaole Sun, Kai Yuan, Wenbin Hu, Chuanyu Yang, and Zhibin Li.
\newblock Learning pregrasp manipulation of objects from ungraspable poses.
\newblock In \emph{2020 IEEE International Conference on Robotics and Automation (ICRA)}, pages 9917--9923. IEEE, 2020.

\bibitem[Sundaralingam et~al.(2023)Sundaralingam, Hari, Fishman, Garrett, Van~Wyk, Blukis, Millane, Oleynikova, Handa, Ramos, et~al.]{sundaralingam2023curobo}
Balakumar Sundaralingam, Siva Kumar~Sastry Hari, Adam Fishman, Caelan Garrett, Karl Van~Wyk, Valts Blukis, Alexander Millane, Helen Oleynikova, Ankur Handa, Fabio Ramos, et~al.
\newblock Curobo: Parallelized collision-free robot motion generation.
\newblock In \emph{IEEE International Conference on Robotics and Automation}, pages 8112--8119, 2023.

\bibitem[Thomason et~al.(2024)Thomason, Kingston, and Kavraki]{thomason2024motions}
Wil Thomason, Zachary Kingston, and Lydia~E Kavraki.
\newblock Motions in microseconds via vectorized sampling-based planning.
\newblock In \emph{IEEE International Conference on Robotics and Automation}, pages 8749--8756, 2024.

\bibitem[Triantafyllidis et~al.(2023)Triantafyllidis, Acero, Liu, and Li]{triantafyllidis2023hybrid}
Eleftherios Triantafyllidis, Fernando Acero, Zhaocheng Liu, and Zhibin Li.
\newblock {Hybrid hierarchical learning for solving complex sequential tasks using the robotic manipulation network ROMAN}.
\newblock \emph{Nature Machine Intelligence}, 5\penalty0 (9):\penalty0 991--1005, 2023.

\bibitem[Wang et~al.(2022)Wang, Hu, Sun, Wang, and Li]{wang2022learning}
Shuaijun Wang, Wenbin Hu, Lining Sun, Xin Wang, and Zhibin Li.
\newblock Learning adaptive grasping from human demonstrations.
\newblock \emph{IEEE/ASME Transactions on Mechatronics}, 27\penalty0 (5):\penalty0 3865--3873, 2022.

\bibitem[Wen et~al.(2025)Wen, Zhu, Li, Tang, Shen, and Feng]{wen2025dexvla}
Junjie Wen, Yichen Zhu, Jinming Li, Zhibin Tang, Chaomin Shen, and Feifei Feng.
\newblock Dexvla: Vision-language model with plug-in diffusion expert for general robot control.
\newblock \emph{arXiv:2502.05855}, 2025.

\bibitem[Xue et~al.(2025)Xue, Ren, Chen, Zhang, Fang, Gu, Xu, and Lu]{xue2025reactive-vtac}
Han Xue, Jieji Ren, Wendi Chen, Gu~Zhang, Yuan Fang, Guoying Gu, Huazhe Xu, and Cewu Lu.
\newblock Reactive diffusion policy: Slow-fast visual-tactile policy learning for contact-rich manipulation.
\newblock \emph{arXiv:2503.02881}, 2025.

\bibitem[Ye et~al.(2025)Ye, Wang, Yuan, Yang, Li, Zhu, Qin, Zou, and Wang]{ye2025dex1b}
Jianglong Ye, Keyi Wang, Chengjing Yuan, Ruihan Yang, Yiquan Li, Jiyue Zhu, Yuzhe Qin, Xueyan Zou, and Xiaolong Wang.
\newblock Dex1b: Learning with 1b demonstrations for dexterous manipulation.
\newblock \emph{arXiv:2506.17198}, 2025.

\bibitem[Zhang et~al.(2025)Zhang, Hao, Cao, Hao, Cui, and Wang]{zhang2025vtla}
Chaofan Zhang, Peng Hao, Xiaoge Cao, Xiaoshuai Hao, Shaowei Cui, and Shuo Wang.
\newblock Vtla: Vision-tactile-language-action model with preference learning for insertion manipulation.
\newblock \emph{arXiv:2505.09577}, 2025.

\bibitem[Zhao et~al.(2023)Zhao, Kumar, Levine, and Finn]{zhao2023learning}
Tony~Z Zhao, Vikash Kumar, Sergey Levine, and Chelsea Finn.
\newblock Learning fine-grained bimanual manipulation with low-cost hardware.
\newblock \emph{arXiv:2304.13705}, 2023.

\bibitem[Zhao et~al.(2025)Zhao, Yu, Jing, and Yang]{zhao2025xrobotoolkit}
Zhigen Zhao, Liuchuan Yu, Ke~Jing, and Ning Yang.
\newblock Xrobotoolkit: A cross-platform framework for robot teleoperation.
\newblock \emph{arXiv:2508.00097}, 2025.

\bibitem[Zhong et~al.(2025)Zhong, Huang, Li, Zhang, Liang, Yang, and Chen]{zhong2025dexgraspvla}
Yifan Zhong, Xuchuan Huang, Ruochong Li, Ceyao Zhang, Yitao Liang, Yaodong Yang, and Yuanpei Chen.
\newblock Dexgraspvla: A vision-language-action framework towards general dexterous grasping.
\newblock \emph{arXiv:2502.20900}, 2025.

\bibitem[Zitkovich et~al.(2023)Zitkovich, Yu, Xu, Xu, Xiao, Xia, Wu, Wohlhart, Welker, Wahid, et~al.]{zitkovich2023rt}
Brianna Zitkovich, Tianhe Yu, Sichun Xu, Peng Xu, Ted Xiao, Fei Xia, Jialin Wu, Paul Wohlhart, Stefan Welker, Ayzaan Wahid, et~al.
\newblock Rt-2: Vision-language-action models transfer web knowledge to robotic control.
\newblock In \emph{Conference on Robot Learning}, pages 2165--2183. PMLR, 2023.

\end{thebibliography}

\end{document}